%% file: FixBPINNs_preprint.tex
\documentclass[twoside]{article}

\usepackage[preprint]{aistats2027}
\usepackage[numbers,sort&compress]{natbib}

\usepackage{subcaption} 
\usepackage{graphicx} 
\usepackage{amsfonts} 
\usepackage[hyphens]{url} 
\usepackage{xcolor}
\usepackage{tabularx}
\usepackage{booktabs}
\usepackage{ragged2e}
\usepackage{mathtools}

\renewcommand{\mid}{\,|\,}

\begin{document}

\twocolumn[

\aistatstitle{Uncovering and Fixing Collider Bias in Bayesian PINNs}

\aistatsauthor{ Michael Obermayr \And Robert Peharz }

\aistatsaddress{TU Graz \And  TU Graz } ]

\begin{abstract}
Bayesian physics-informed neural networks (B-PINNs) are a popular framework for parameter and state inference from sparse or noisy observations. 
They are commonly formulated via a \emph{collider structure}, in which physical and trajectory parameters are assumed to be a priori independent and become coupled through virtual likelihoods on differential-equation residuals that enforce physical consistency. 
We show that this modeling choice can induce severe systematic bias in the posterior over physical parameters: even when the prior is favorably centered on the ground-truth parameters, the resulting posterior can drift away and concentrate far from them. 
As a remedy, we advocate a hierarchical \emph{chain model} in which physics generates trajectories, which in turn generate observations. 
The chain model does not suffer from this posterior bias, but it poses a harder, so-called doubly intractable, inference problem due to a physics-dependent normalization constant. 
This challenge can be resolved by discretizing the underlying stochastic dynamics, after which the chain posterior can be sampled exactly with particle MCMC.
We identify two distinct mechanisms characterizing the collider bias, derive analytical approximations of their magnitudes, and establish diagnostic criteria for predicting when standard B-PINNs remain reliable. 
Experiments confirm the predicted bias and show that the chain formulation successfully avoids it.
\end{abstract}

\section{Introduction}
Inferring unknown physical parameters from dynamical systems is a fundamental challenge across science and engineering.
Physics-informed neural networks (PINNs)~\citep{Raissi19_PINN} tackle this inverse problem by training neural trajectory surrogates to jointly minimize data mismatch and physical law violations.
Under limited and noisy data, a Bayesian treatment naturally quantifies uncertainty over trajectories and parameters.
Yet, their assumed dependency structure requires careful consideration, which is the focus of this paper.

\begin{figure}
    \centering
    \begin{subfigure}[c]{0.45\columnwidth}
        \includegraphics[width=\textwidth]{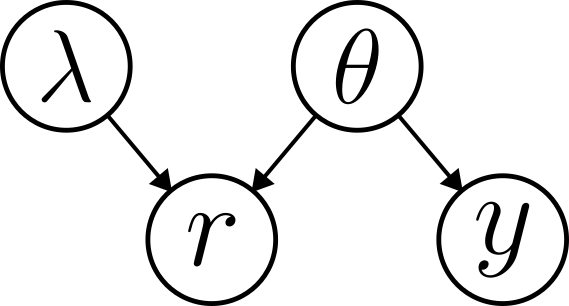}
        \subcaption{Collider (B-PINN)}
        \label{subfig:Collider-Model}
    \end{subfigure}~~~~~
    \begin{subfigure}[c]{0.45\columnwidth}
        \includegraphics[width=\textwidth]{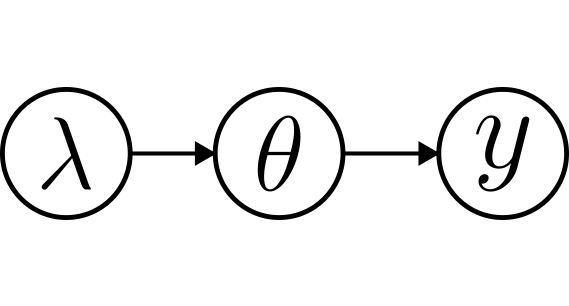}
        \subcaption{Chain (ours)}
        \label{subfig:Chain-Model}
    \end{subfigure}
    \caption{Modeling choices for physics-informed Bayesian inference. \textbf{(a) Collider (B-PINN):} 
    Physics parameters $\lambda$ and trajectory parameters $\theta$ are a priori independent, interacting only when conditioned on equation residuals $r$. \textbf{(b) Chain (ours):} Physics $\lambda$ directly generates trajectories $\theta$, which in turn generate data $y$.}
    \label{fig:GraphicalModels}
\end{figure}

\begin{figure*}
    \centering
    \includegraphics[width=\textwidth]{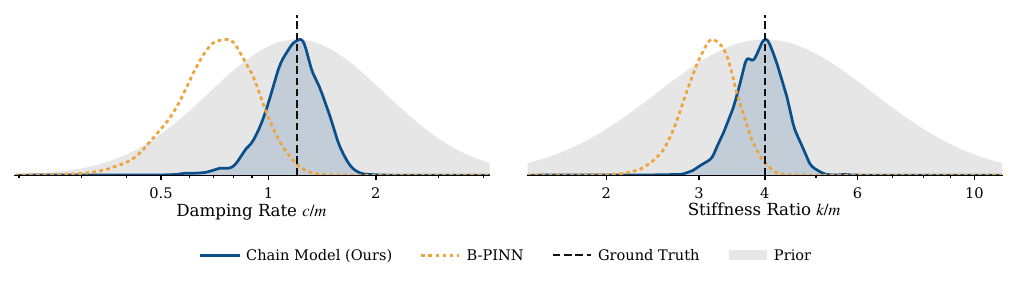}
    \caption{\textbf{Posterior marginals for the damped linear oscillator.}
    System governed by $\ddot{x} = -\frac{c}{m}\dot{x} - \frac{k}{m}x + \sigma_r \xi(t)$ with white noise $\xi(t)$, mass $m$, damping $c$ and spring constant $k$, ground-truth parameters $\{m, c, k\} = \{1.0, 1.2, 4.0\}$, residual scale fixed at $\sigma_r = 0.3$, and 40 noisy observations ($\sigma_\text{obs} = 0.02$).
    Posteriors are sampled via NUTS (24,000 samples) for B-PINN and PMMH~\citep{Andrieu2010_PMMH} (14,000 samples) for our proposed chain model.
    Although all three physical parameters are inferred jointly, only the ratios $c/m$ and $k/m$ are identifiable; we therefore display their corresponding marginal posteriors.
    Despite initialization at the ground truth and identical priors centered at the ground truth, only the chain model accurately recovers the true values.
    In contrast, B-PINN exhibits a systematic collider bias, underestimating dissipation ($c/m$) and stiffness ($k/m$).
    The bias is reproducible over resampled datasets.
    Full experimental details are provided in Section~\ref{sec:Experiments}.}
    \label{fig:Exp_Osc_Posteriors}
\end{figure*}

\emph{Bayesian PINNs} (B-PINNs)~\citep{Yang2021_BPINN} assume the factorization
\begin{equation} \label{eq:BPINN-Joint}
    p_{\mathrm{collider}}(\lambda, \theta, r, y) 
    = \underbrace{p(\lambda)}_{\mathclap{\substack{\text{\tiny Physics}\\[-1pt]\text{\tiny Prior}}}} 
    \, \underbrace{p(\theta)}_{\mathclap{\substack{\text{\tiny Traj.}\\[-1pt]\text{\tiny Prior}}}}
    \, \underbrace{p(r \mid \lambda, \theta)}_{\mathclap{\substack{\text{\tiny Residual}\\[-1pt]\text{\tiny Likelihood}}}} 
    \, \underbrace{p(y \mid \theta)}_{\mathclap{\substack{\text{\tiny Observ.}\\[-1pt]\text{\tiny Likelihood}}}}. 
\end{equation}
Here $p(\lambda)$ and $p(\theta)$ are the priors over physics and trajectory parameters, respectively, and $p(y\mid\theta)$ is the observation likelihood.
Residuals $r$ serve as ``virtual observations'' of zero measuring physics violations at collocation points, typically under a narrow zero-mean Gaussian likelihood $p(r\mid\lambda,\theta)$.
Thus, $\lambda$ and $\theta$ are a priori independent and coupled only upon conditioning on $r=\mathbf{0}$.
In graphical-model terms~\citep{koller2009probabilistic}, this forms a \emph{collider} at $r$ (Fig.~\ref{subfig:Collider-Model}), a structure widely adopted in the literature~\citep{Sun2020_BPINN, Linka2022_BPINN, Stock2023_BPINN, Pensoneault2024}.

Alternatively, one might use a hierarchical model, in which physics $\lambda$ is equipped with a prior, trajectories $\theta$ depend on $\lambda$ and observations $y$ depend on $\theta$:
\begin{equation} \label{eq:Chain-Joint} 
  p_{\mathrm{chain}}(\lambda, \theta, y) 
  \propto ~\underbrace{p(\lambda)}_{\mathclap{\substack{\text{\tiny Physics}\\[-1pt]\text{\tiny Prior}}}} 
  ~~ \underbrace{p(\theta \mid \lambda)}_{\mathclap{\substack{\text{\tiny Trajectory}\\[-1pt]\text{\tiny Likelihood}}}} 
  ~~ \underbrace{p(y \mid \theta)}_{\mathclap{\substack{\text{\tiny Observation}\\[-1pt]\text{\tiny Likelihood}}}}.
\end{equation}
Here the dependency structure between $\lambda$ and $\theta$ forms a \emph{chain} (head-to-tail node) \citep{koller2009probabilistic}, as illustrated in Fig.~\ref{subfig:Chain-Model}.
Unlike the collider \eqref{eq:BPINN-Joint}, $\theta$ depends directly on $\lambda$ through the \emph{trajectory likelihood} $p(\theta \mid \lambda)$ rather than indirectly through residual likelihood $p(r \mid \lambda, \theta)$.

\textbf{Which of the two formulations is preferable?}
The collider offers practical convenience: its factors are straightforward Gaussian priors and likelihoods.
By contrast, specifying $p(\theta \mid \lambda)$ in the chain is less straightforward; it requires describing a distribution over trajectory parameters $\theta$ that may vary substantially with the physics $\lambda$.
An intuitive construction is to start from the prior $p(\theta)$, similar as in the collider formulation, reweight it according to the exponential of the negative squared residual error, and renormalize the result.
As detailed in Section~\ref{sec:Background}, this renders the chain model~\eqref{eq:Chain-Joint} nearly identical to the collider model~\eqref{eq:BPINN-Joint}, up to a \emph{physics-dependent} normalization constant $Z_\lambda$ whose calculation requires an intractable marginalization over $\theta$.

\textbf{Despite the computational simplicity of the collider, the chain should be preferred.}
Conceptually, the residuals $r$ in the collider formulation act merely as auxiliary variables to couple physics $\lambda$ and trajectories $\theta$, rather than quantities of substantive interest. 
The chain avoids this artificial construction by modeling trajectory dependence directly via $p(\theta \mid \lambda)$. 
Furthermore, forward-sampling the collider yields an implausible generative process—typically generating incompatible pairs of $\lambda$ and $\theta$ that demand unlikely residuals. 
By contrast, the chain follows the natural hierarchy in which physics $\lambda$ generates trajectories $\theta$, which in turn generate observations $y$.

\textbf{Crucially, the collider systematically biases the posterior over physical parameters.}
To demonstrate this, we sample synthetic data from a damped linear oscillator with known ground truth, center prior modes at the true values, and sample the collider posterior with NUTS~\citep{Hoffman2014_NUTS}, using Fourier features to prevent spectral bias~\citep{Wang2021_FourierFeature}.
Despite these favorable conditions, the collider marginals \emph{drift substantially from the ground truth} (Fig.~\ref{fig:Exp_Osc_Posteriors}), by about two posterior standard deviations in $c/m$.
This is alarming, as the collider model can return confidently wrong estimates of the underlying physics.
By contrast, our proposed chain inference exhibits the expected behavior: its posterior concentrates at the ground truth.

As shown in Section~\ref{sec:Background}, this bias stems directly from the physics-dependent normalization constant $Z_\lambda$, which systematically favors less restrictive physics in the collider setting. 
While the chain formulation eliminates this bias, it yields a challenging, \emph{doubly-intractable problem}~\citep{Murray2006}. 
Despite the prevalence of collider-based approaches in physics-informed neural inference, this structural bias remains largely unaddressed.
To our knowledge, the only exceptions are Alberts, Hao, and Bilionis~\citep{Bilionis1, Bilionis2, Bilionis3}, who employ a similar chain structure and trace the posterior bias to $Z_\lambda$; however, they address the doubly-intractable problem via computationally demanding nested inference without further analyzing the bias.
Still, their formulation has gained little traction in the community and the collider formulation remains the default in B-PINN practice~\citep{Moon2025_ThermalBPINN, Shukla2026_TurbulentUQ}.


\section{Problem Formulation and Probabilistic Modeling} \label{sec:Background}
 
We consider parameterized second-order ordinary differential equations (ODEs) governing a trajectory $x(t) \in \mathbb{R}^d$:
\begin{equation} \label{eq:ODE}
    M_\lambda(x, \dot{x}) \, \ddot{x} = F_\lambda(x, \dot{x}, t),
\end{equation}
where $\dot{x}$ and $\ddot{x}$ denote the first and second time derivatives, $M_\lambda(x, \dot{x}) \in \mathbb{R}^{d \times d}$ is an invertible generalized mass matrix, and $F_\lambda(x, \dot{x}, t) \in \mathbb{R}^d$ is a force field driving the system, both parameterized by unknown physical parameters $\lambda \in \mathbb{R}^p$.
Equation~\eqref{eq:ODE} is the general form of a broad class of second-order dynamical systems arising in classical mechanics, structural dynamics, and robotics~\citep{goldstein2002classical, murray1994mathematical}.
 
In practice, the trajectory $x(t)$ is rarely observed directly and is therefore treated as latent.
Instead, we only have access to sparse, noisy observations $y = \{y_k\}_{k=1}^K$ at times $t_k$, governed by a Gaussian measurement model
\begin{equation} \label{eq:obs-model}
    y_k = x(t_k) + \epsilon_k, \quad \epsilon_k \sim \mathcal{N}(0, \Sigma_{\mathrm{obs}}),
\end{equation}
with known observation covariance $\Sigma_{\mathrm{obs}}$.
 
Beyond the measurement, the dynamics themselves are typically uncertain: simplifications and unmodeled degrees of freedom inevitably perturb the ideal dynamics~\eqref{eq:ODE}.
We account for these discrepancies by introducing process noise on the velocity $v := \dot{x}$.
With this, the phase-space state $z := (x^\top, v^\top)^\top \in \mathbb{R}^{2d}$ evolves according to an It\^o stochastic differential equation (SDE)
\begin{equation} \label{eq:Ito-SDE}
    \mathrm{d}z_t = \underbrace{\begin{bmatrix} v_t \\ a_\lambda(z_t, t) \end{bmatrix}}_{=:\, f_\lambda(z_t, t)} \mathrm{d}t
    + \begin{bmatrix} 0 \\ A_\lambda(z_t)^{-1} \Sigma_r^{1/2} \end{bmatrix} \mathrm{d}W_t,
\end{equation}
driven by standard Brownian motion $W_t \in \mathbb{R}^d$.
Here, $a_\lambda(z, t) := M_\lambda(x, v)^{-1} F_\lambda(x, v, t)$ is the acceleration field and $f_\lambda$ the drift.
The physical disturbances are modeled as constant white noise with covariance $\Sigma_r$, and the invertible, possibly state- and parameter-dependent matrix $A_\lambda(z) \in \mathbb{R}^{d \times d}$ maps them into velocity increments.

The choice of $A_\lambda$ determines how process noise enters the dynamics: setting $A_\lambda = I_d$ injects white noise directly into the accelerations, whereas setting $A_\lambda = M_\lambda$ models the perturbations as stochastic forces. 
The latter is more realistic for many mechanical systems, where physical disturbances act as fluctuating generalized forces or torques. 
As we show in Section~\ref{sec:Theory}, parameter dependence in $A_\lambda$ is one of the factors contributing to collider bias in the inferred physical parameters.

With process noise present ($\Sigma_r \succ 0$), particular physics parameters $\lambda$ induce a distribution $p(x \mid \lambda)$ over trajectories rather than a single deterministic trajectory. 
Our primary goal is to infer the marginal posterior distribution over the physical parameters given sparse observations of the trajectory:
\begin{equation} 
p(\lambda \mid y) \propto p(\lambda) \int p(y \mid x) \, p(x \mid \lambda) \, \mathrm{d}x.
\end{equation}
Following standard B-PINN practice, we fix the residual scale $\Sigma_r$ as a hyperparameter rather than inferring it alongside $\lambda$~\citep{Yang2021_BPINN, Linka2022_BPINN}.
 
\subsection{Physics-Informed Residuals and Energy} \label{sec:Background-Residuals}

To connect the trajectory distribution induced by the stochastic dynamics~\eqref{eq:Ito-SDE} with physics-informed inference, we express deviations from the governing dynamics through a residual energy. 
In physics-informed machine learning, the continuous trajectory is represented by a differentiable surrogate $x_\theta(t)$ parameterized by $\theta$ (e.g., neural network weights). 
At a set of collocation points $t_1 < \dots < t_N$, we define the physics residual
\begin{equation} \label{eq:Residual-Unified}
r_n(\theta, \lambda) := A_\lambda(z_\theta(t_n))
\left(
\ddot{x}_\theta(t_n) - a_\lambda(z_\theta(t_n), t_n)
\right),
\end{equation}
where $z_\theta := (x_\theta^\top, \dot{x}_\theta^\top)^\top$.
To relate these pointwise residuals to the process-noise model, suppose the collocation points are uniformly spaced with step size $h$. 
Over an interval of width $h$, the Brownian increment driving~\eqref{eq:Ito-SDE} has covariance $h\Sigma_r$. 
Equivalently, the corresponding average stochastic forcing over that interval has covariance $\Sigma_r/h$. 
This motivates modeling the residuals at the collocation points as Gaussian deviations with covariance $\Sigma_r/h$, yielding the quadratic residual energy
\begin{equation} \label{eq:EnergyFunctional}
E[\theta, \lambda]
=
\frac{h}{2}
\sum_{n=1}^N
r_n(\theta, \lambda)^\top
\Sigma_r^{-1}
r_n(\theta, \lambda).
\end{equation}

Thus, the stochastic dynamics induce a natural quadratic energy measuring the compatibility between a trajectory $x_\theta$ and physical parameters $\lambda$. 
This residual energy is the common ingredient underlying the probabilistic formulations considered next.


\subsection{Probabilistic Models: Collider vs. Chain} \label{sec:Background-Models}
We can now define the two probabilistic models---collider \eqref{eq:BPINN-Joint} and chain \eqref{eq:Chain-Joint}---in a parallel manner and using similar components. 
Both factorizations share the same prior $p(\lambda)$ over physical parameters and the same observational likelihood,
\begin{equation} \label{eq:Observation-Likelihood}
p(y \mid \theta) = \prod_{k=1}^K \mathcal{N}\left(y_k \mid x_\theta(t_k), \Sigma_{\mathrm{obs}}\right).
\end{equation}
The key difference between them is how the residual energy~\eqref{eq:EnergyFunctional} is used to couple the trajectory parameters $\theta$ and the physical parameters $\lambda$.

\paragraph{Collider formulation.}
Standard Bayesian PINNs~\citep{Yang2021_BPINN, Sun2020_BPINN, Linka2022_BPINN, Stock2023_BPINN} place an independent prior $p(\theta)$ on the trajectory parameters and enforce the physical law through auxiliary residuals treated as pseudo-observations equipped with a narrow zero-mean Gaussian:
\begin{equation} \label{eq:BPINN-Likelihood}
\begin{split}
    p(r = \mathbf{0} \mid \lambda, \theta) & = \prod_{n=1}^N \mathcal{N}\left(\mathbf{0} \middle| r_n(\theta, \lambda), \tfrac{1}{h} \Sigma_r\right) \\
    &= \frac{1}{Z_r} \exp\left(-E[\theta, \lambda]\right),
\end{split}
\end{equation}
where the Gaussian normalization constant $Z_r = \det(2\pi \Sigma_r / h)^{N/2}$ depends on $\Sigma_r$, but is invariant to $\lambda$ and $\theta$.
Because $\lambda$ and $\theta$ are assumed a priori independent and interact only via conditioning on $r = \mathbf{0}$, the residual acts as a collider node in the underlying directed acyclic graph (Fig.~\ref{subfig:Collider-Model}).

\paragraph{Chain formulation.}
In contrast, the hierarchical chain formulation follows the generative direction $\lambda \to \theta \to y$ (Fig.~\ref{subfig:Chain-Model}) by defining a conditional distribution over trajectories,
\begin{equation} \label{eq:Trajectory-Likelihood}
p(\theta \mid \lambda) = \frac{1}{Z_\lambda} \exp\left(-E[\theta, \lambda]\right) p(\theta).
\end{equation}
Here, $p(\theta)$ is interpreted as a reference density over the trajectory parametrization, which can encode preferences for certain trajectories. 
Unlike in the collider formulation, it is not a prior over $\theta$; instead, the exponential factor tilts this reference density toward trajectories that are compatible with the $\lambda$. 
Normalizing the resulting conditional distribution requires the normalization constant
\begin{equation} \label{eq:Z-Integral}
Z_\lambda = \int \exp\left(-E[\theta, \lambda]\right) p(\theta) \, \mathrm{d}\theta,
\end{equation}
which generally \emph{depends on the physical parameters $\lambda$.}

\paragraph{Model Difference and The Collider Bias.}
The two formulations differ only through their normalization factors and we can write:
\begin{equation} \label{eq:Model_Difference}
p_{\mathrm{collider}}(\lambda, \theta, r = \mathbf{0}, y)
=
\frac{Z_\lambda}{Z_r}
p_{\mathrm{chain}}(\lambda, \theta, y).
\end{equation}
Since $Z_r$ is constant with respect to $\lambda$, this difference carries directly to the posterior over the physical parameters:
\begin{equation} \label{eq:Delta-Definition}
\Delta(\lambda):= 
\log
\frac{
p_{\mathrm{collider}}(\lambda \mid y, r=\mathbf{0})
}{
p_{\mathrm{chain}}(\lambda \mid y)
}
=
\log Z_\lambda + C,
\end{equation}
where we call the model difference $\Delta(\lambda)$ the \emph{collider bias}.
It is the mechanism underlying the posterior drift observed in Fig.~\ref{fig:Exp_Osc_Posteriors}.
The term \emph{bias} is deliberate: $\Delta$ displaces the posterior from the data-generating parameters, so that the collider's credible intervals miss the ground truth, not merely the chain posterior (Section~\ref{sec:Experiments}).
The partition function $Z_\lambda$ measures the amount of trajectory space, weighted by the reference density $p(\theta)$, that is compatible with the dynamics under $\lambda$. 
The collider formulation therefore reweights the desired posterior by $Z_\lambda$, favoring $\lambda$ that admit a larger volume of low-residual trajectories, that is, less restrictive physics. 

\section{Analysis of the Collider Bias} \label{sec:Theory}
For B-PINNs with a neural trajectory surrogate, the collider bias $\Delta(\lambda)$ decomposes into two distinct mechanisms. 
In Section~\ref{sec:Theory-Exact}, we show that adopting a Markovian state-space latent model eliminates one mechanism and renders the other analytically tractable.

\paragraph{Laplace approximation of the collider bias.}
Because the partition function~\eqref{eq:Z-Integral} is analytically intractable for neural surrogates, we place a Gaussian prior $p(\theta) = \mathcal{N}(0, P^{-1})$ on the weights and apply a Laplace expansion around the data-free mode $\theta^*(\lambda)$.
The collider bias~\eqref{eq:Delta-Definition} evaluates to
\begin{equation} \label{eq:Delta-Expansion}
    \Delta(\lambda) \approx -\frac{1}{2} \log\det\left( G_\lambda + P \right) + C,
\end{equation}
where $G_\lambda$ is the Gauss--Newton curvature of the residual energy~\eqref{eq:EnergyFunctional} at $\theta^*$. 
The residual constrains $n_{\mathrm{eff}} := \operatorname{tr}\!\left[ G_\lambda ( G_\lambda + P )^{-1} \right] \le \operatorname{rank} G_\lambda \le \min(Nd, \dim\theta)$ directions. 
Evaluating this curvature along the nominal trajectory $z(t) := z_{\theta^*}(t)$ over $t \in [0, \mathcal{T}]$ decomposes~\eqref{eq:Delta-Expansion} into:
\begin{equation} \label{eq:TwoBiasesDecomposition}
\begin{split}
    \Delta(\lambda) \approx
    & \underbrace{-\frac{n_{\mathrm{eff}}}{Nd} \sum_{n=1}^N \log \lvert\det A_\lambda(z(t_n))\rvert}_{\text{\textbf{(1) Residual metric bias}}} \\
    &\underbrace{- \frac{1}{2} \int_0^{\mathcal{T}} \sum_{k} \bigl\lvert\operatorname{Re} s_k(t)\bigr\rvert \, \mathrm{d}t}_{\text{\textbf{(2) Dissipation bias}}} \; + \; C,
\end{split}
\end{equation}
where $\{s_k(t)\}$ are the eigenvalues of the drift Jacobian $Df_\lambda$ along $z(t)$ (full derivation in Appendix~\ref{sec:Tilt-Derivation}).
Eq.~\eqref{eq:TwoBiasesDecomposition} explains the origin and direction of the bias.
 
\paragraph{Derivation Assumptions.}
This decomposition relies on four simplifying assumptions (Appendix~\ref{sec:Tilt-Derivation}):
(i)~the surrogate closely satisfies the ODE at $\theta^*$, with the prior penalty depending only weakly on $\lambda$;
(ii)~$G_\lambda$ is evaluated along a single nominal trajectory;
(iii)~the residual loss dominates the prior along constrained directions ($n_{\mathrm{eff}} \approx \operatorname{rank} G_\lambda$ locally constant); and
(iv)~the residual metric and drift decouple additively.
We also treat $A_\lambda$ and $Df_\lambda$ as piecewise constant in time, and assume the surrogate bandwidth $\Omega$ exceeds the system's characteristic dynamical rates.

\paragraph{Mechanism 1: Residual metric bias.}
Because $Z_r$ normalizes over residual space, it omits the Jacobian determinant $\lvert\det A_\lambda\rvert$ of the acceleration-to-residual mapping, whereas $Z_\lambda$ accounts for it along all $n_{\mathrm{eff}}$ constrained directions.
Consequently, acceleration residuals ($A_\lambda = I_d$) avoid this bias, whereas force residuals ($A_\lambda = M_\lambda$) add $-\frac{n_{\mathrm{eff}}}{Nd}\sum_{n=1}^N \log \lvert\det M_\lambda(z(t_n))\rvert$, shifting posterior mass toward vanishing mass determinants.
The residual metric is dictated by where the noise enters the dynamics (Section~\ref{sec:Background-Residuals}); switching to acceleration residuals for convenience changes the noise model and is therefore not a free choice.

This pathology worsens along drift-invariant parameter directions where jointly scaling masses and forces preserves drift.
Because the bias grows with $n_{\mathrm{eff}}$ while the data only constrain scale via finite observations, the artifact can overwhelm the likelihood and collapse the posterior toward vanishing masses (Appendix~\ref{sec:Ray-Collapse}).

\paragraph{Mechanism 2: Dissipation bias.}
Lacking Markovian temporal structure, the neural surrogate couples perturbations through the operator $\frac{\mathrm{d}}{\mathrm{d}t} - Df_\lambda$ nonlocally over $[0, \mathcal{T}]$, generating the second term in~\eqref{eq:TwoBiasesDecomposition}.

When all perturbations decay ($\operatorname{Re} s_k \le 0$), the spectral sum reduces to phase-space divergence: $\sum_k \lvert\operatorname{Re} s_k\rvert = -\nabla \cdot f_\lambda = -\operatorname{tr}(\partial a_\lambda / \partial v)$. 
The dissipation bias simplifies to
\begin{equation} \label{eq:NonlocalBias}
    \Delta_{\mathrm{diss}}(\lambda) = \frac{1}{2} \int_0^{\mathcal{T}} \nabla \cdot f_\lambda(z(t), t) \, \mathrm{d}t + \mathcal{O}(\mathcal{T}/\Omega) + C,
\end{equation}
whose leading term matches the Onsager--Machlup path-action correction~\citep{OnsagerMachlup1953}.
In dissipative systems ($\nabla \cdot f_\lambda < 0$), $e^{\Delta_{\mathrm{diss}}}$ spuriously penalizes dissipation, underestimating friction, damping, or drag.
Crucially, this penalty scales as $\Theta(\mathcal{T})$, so that the B-PINN's error grows with the record length relative to its own posterior width (Section~\ref{sec:Exp-Grid}).
In the presence of growing modes, it analogously penalizes stronger instability (Appendix~\ref{sec:Tilt-Derivation}).


\section{Tractable Inference in the Chain Model} \label{sec:Method}
With a global surrogate, the chain model yields a doubly intractable posterior~\citep{Murray2006}, usually requiring expensive inner-loop re-estimation of $Z_\lambda$ via variational inference or Langevin dynamics that often diverges (Appendix~\ref{sec:BilionisComparison}).
Instead of attempting global integration, we discretize the generative SDE~\citep{Sarkka2019}. 
This factorizes $Z_\lambda$ into analytic per-step normalizers~\eqref{eq:Z-StateSpace}, enabling asymptotically exact inference via particle MCMC.

\subsection{The Chain Model as a State-Space Model} \label{sec:StateSpaceModel}
Discretizing $[0, \mathcal{T}]$ into $T$ intervals of width $h$ defines states $z_i = (x_i^\top, v_i^\top)^\top$ at $t_i = ih$ and applying a semi-implicit Euler--Maruyama scheme to~\eqref{eq:Ito-SDE} yields:
\begin{equation} \label{eq:EM}
\begin{split}
    v_i &= v_{i-1} + h \, a_\lambda(z_{i-1}, t_{i-1}) + \sqrt{h} \, A_\lambda(z_{i-1})^{-1} \Sigma_r^{1/2} \xi_i, \\
    x_i &= x_{i-1} + h \, v_i,
\end{split}
\end{equation}
with independent increments $\xi_i \sim \mathcal{N}(0, I_d)$.
This discretization induces a Markovian trajectory prior over $\theta = z_{0:T}$:
\begin{equation} \label{eq:Chain-TransitionKernel}
\begin{split}
    p(z_{0:T} \mid \lambda) &= p(z_0) \prod_{i=1}^T K_\lambda(z_i \mid z_{i-1}), \\
    K_\lambda(z_i \mid z_{i-1}) &= \delta\!\left(x_i - x_{i-1} - h v_i\right) \\
    &\quad \cdot \mathcal{N}\!\left(v_i \,\middle|\, \bar{v}_i, \; h \, A_\lambda^{-1} \Sigma_r A_\lambda^{-\top}\right),
\end{split}
\end{equation}
where $\bar{v}_i := v_{i-1} + h \, a_\lambda(z_{i-1}, t_{i-1})$ and $A_\lambda$ is evaluated at $z_{i-1}$.
Unlike~\eqref{eq:Trajectory-Likelihood}, this prior admits analytic, per-step normalization.
Combined with the prior $p(\lambda)$ and observations at $\{t_k\} \subseteq \{t_i\}_{i=1}^T$, this completes the chain model~\eqref{eq:Chain-Joint} over $(\lambda, z_{0:T})$ to target the exact marginal posterior $p(\lambda \mid y)$.

\subsection{Exact Collider Bias on the State-Space Latent} \label{sec:Theory-Exact} 
On the Markovian latent state $\theta = z_{0:T}$, the collider bias can be evaluated in closed form.
The velocity update in~\eqref{eq:EM} corresponds to the discrete residual
\begin{equation} \label{eq:Residual-Discrete}
    r_i = A_\lambda(z_{i-1}) \left( \frac{v_i - v_{i-1}}{h} - a_\lambda(z_{i-1}, t_{i-1}) \right),
\end{equation}
making the transition kernel~\eqref{eq:Chain-TransitionKernel} proportional to $\exp(-\frac{h}{2} r_i^\top \Sigma_r^{-1} r_i)$, which matches the summand of~\eqref{eq:EnergyFunctional} for $N = T$.
Because $r_i$ is affine in $v_i$, each one-step transition normalizes analytically:
\begin{equation} \label{eq:Z-StateSpace}
\begin{split}
    Z_\lambda(z_{i-1}) &:= \int \exp\left( -\frac{h}{2} r_i^\top \Sigma_r^{-1} r_i \right) \mathrm{d}v_i \\
    &\phantom{:}= \frac{(2\pi h)^{d/2} \det(\Sigma_r)^{1/2}}{\lvert\det A_\lambda(z_{i-1})\rvert}.
\end{split}
\end{equation}
The drift $a_\lambda$ acts merely as a mean shift and drops out of the integral, causing the dissipation bias to vanish.
Substituting $\prod_{i=1}^T Z_\lambda(z_{i-1})$ for $Z_\lambda$ in~\eqref{eq:Model_Difference} and subtracting $\log Z_r = \frac{T}{2} \log\det(2\pi\Sigma_r / h)$ gives the exact collider bias:
\begin{equation} \label{eq:Metric-Tilt-Exact}
    \Delta_{\mathrm{Markov}}(\lambda, z_{0:T-1}) = -\sum_{i=1}^T \log \lvert\det A_\lambda(z_{i-1})\rvert + C.
\end{equation}
Only the residual metric bias remains: under acceleration residuals ($A_\lambda = I_d$), $\Delta_{\mathrm{Markov}}$ is independent of $\lambda$ and both posteriors coincide; under force residuals ($A_\lambda = M_\lambda$), $\Delta_{\mathrm{Markov}}$ biases the collider posterior toward vanishing mass determinants.

\subsection{Exact Parameter Inference via Particle MCMC} \label{sec:InferenceScheme}
For nonlinear dynamics, the marginal likelihood $p(y \mid \lambda) = \int p(z_{0:T} \mid \lambda) \, p(y \mid z_{0:T}) \, \mathrm{d}z_{0:T}$ lacks an analytic form.
Rather than approximating it variationally~\citep{Archambeau2007_VI}, we exploit its Markovian structure, which admits sequential Monte Carlo estimation via a bootstrap particle filter~\citep{Gordon1993_Bootstrap}.
We propagate $P$ particles forward under~\eqref{eq:EM}, reweight by $\mathcal{N}(y_k \mid x_{t_k}, \Sigma_{\mathrm{obs}})$ at observation times $t_k$, and adaptively resample~\citep{LiuChen1998_SMC}.
The resulting estimator $\hat{p}(y \mid \lambda)$ costs $\mathcal{O}(P T)$ and is strictly unbiased.
By the pseudo-marginal property~\citep{Andrieu2010_PMMH}, substituting $\hat{p}(y \mid \lambda)$ into the Metropolis--Hastings acceptance ratio guarantees Particle Marginal Metropolis--Hastings (PMMH) targets the exact posterior $p(\lambda \mid y)$ for any $P \ge 1$, where $P$ governs only chain mixing.
Sampling a trajectory alongside each accepted $\lambda'$ additionally yields exact joint samples from $p(\lambda, z_{0:T} \mid y)$.


\section{Experiments} \label{sec:Experiments}
 
We test the analysis of Section~\ref{sec:Theory} on three systems that isolate its bias mechanisms, and compare our inference scheme with nested inference.


\begin{table*}[t]
    \caption{\textbf{The chain model recovers the truth in every system; the collider fails where Section~\ref{sec:Theory} predicts.}
    $\log\alpha$: posterior mean of the log scale factor along the weakly identified drift-invariant direction (Eq.~\eqref{eq:Ray-alpha}; correct: $\approx 0$). 
    Other columns: posterior mean of an identified quantity divided by its true value (correct: $1.00$). 
    \textbf{Bold}: off by more than a factor of two. $\dagger$: chains did not converge or stalled; median of the draws, for both columns of the system. $\ddagger$: exact Kalman likelihood, since PMMH cannot reach this posterior. 
    All parameters in Appendix~\ref{sec:Exp-FullGrid}.}
    \label{tab:grid}
    \vskip 0.05in
    \begin{center}
    \begin{small}
    \setlength{\tabcolsep}{3pt}
    \begin{tabular*}{\textwidth}{@{\extracolsep{\fill}} l cc cc cc @{}}
    \toprule
    & \multicolumn{2}{c}{\textbf{Oscillator}} & \multicolumn{2}{c}{\textbf{Double Pendulum}} & \multicolumn{2}{c}{\textbf{Relativistic Particle}} \\
    \cmidrule(lr){2-3} \cmidrule(lr){4-5} \cmidrule(lr){6-7}
    \textbf{Model (latent)} & $\log\alpha$ & $c/m$ & $\log\alpha$ & $m_2/m_1$ & $\log\alpha$ & $c$ \\
    \midrule
    \multicolumn{7}{@{}l}{\textit{\textbf{Acceleration residuals} ($A_\lambda = I_d$): metric bias inactive}} \\
    \addlinespace[2pt]
    Chain (state space)               & $-0.01$ & 0.99 & $-0.03$ & 0.99 & $-0.02$ & 0.89 \\
    Collider (state space)            & $-0.01$ & 0.99 & $-0.03$ & 0.99 & $-0.02$ & 0.89 \\
    Chain (Fourier MLP)               & $-0.00$ & 1.00 & --- & --- & --- & --- \\
    Collider B-PINN (Fourier MLP)     & $-0.00$ & 0.61 & $-0.03^{\dagger}$ & 0.94 & $+0.00$ & 1.78 \\
    \midrule
    \multicolumn{7}{@{}l}{\textit{\textbf{Force / torque residuals} ($A_\lambda = M_\lambda$): metric bias active}} \\
    \addlinespace[2pt]
    Chain (state space)               & $-0.01$ & 1.01 & $+0.04$ & 1.00 & $-0.18$ & 0.89 \\
    Collider (state space)            & $\mathbf{-42.42}^{\ddagger}$ & \textbf{27} & $\mathbf{-2.37}^{\dagger}$ & \textbf{0.30} & $\mathbf{-8.17}^{\dagger}$ & $\mathbf{4.1{\cdot}10^{3}}$ \\
    Chain (Fourier MLP)               & $+0.29$ & 0.74 & --- & --- & --- & --- \\
    Collider B-PINN (Fourier MLP)     & $\mathbf{-0.93}$ & \textbf{2.05} & $\mathbf{-2.79}$ & 0.54 & $\mathbf{-1.88}$ & 1.61 \\
    \bottomrule
    \end{tabular*}
    \end{small}
    \end{center}
    \vskip -0.1in
\end{table*}

\subsection{Systems and Setup} \label{sec:Exp-Systems}

Data are generated from the SDE~\eqref{eq:Ito-SDE} with isotropic process noise $\Sigma_r = \sigma_r^2 I_d$ fixed at ground truth; physics priors are log-normal with medians at the truth. Jointly scaling masses and forces by $\alpha$ leaves the drift invariant; this \emph{drift-invariant direction} is only weakly identifiable via process noise.

\textbf{Damped oscillator} ($d=1$): mass $m$, stiffness $k$, linear damping $c$, drift $m\ddot{x} = -c\dot{x} - kx$, truth $(m, c, k) = (1, 1.2, 4)$, and $\sigma_r = 0.3$. Because only $c/m$ and $k/m$ enter the drift, the invariant direction is $(m,c,k) \to \alpha(m,c,k)$. We observe position $K = 40$ times over $\mathcal{T} = 24$ ($\sigma_{\mathrm{obs}} = 0.02$). Dissipation bias is active ($\nabla\!\cdot\! f_\lambda = -c/m < 0$).

\textbf{Double pendulum} ($d=2$): masses $m_1, m_2$ on massless rods of lengths $\ell_1, \ell_2$, with vertical angles $x = (\vartheta_1, \vartheta_2)$. The drift is $M_\lambda(x)\,\ddot{x} = F_\lambda(x, \dot{x})$ with state-dependent mass matrix $M_\lambda$ and forces $F_\lambda$:
\begin{gather*}
    M_\lambda = \begin{bmatrix} (m_1{+}m_2)\ell_1^2 & \mu\cos\delta \\ \mu\cos\delta & m_2\ell_2^2 \end{bmatrix}, \\
    F_\lambda = -\begin{bmatrix} \mu\dot\vartheta_2^2\sin\delta + (m_1{+}m_2)g\ell_1\sin\vartheta_1 \\ -\mu\dot\vartheta_1^2\sin\delta + m_2 g\ell_2\sin\vartheta_2 \end{bmatrix},
\end{gather*}
with $\delta := \vartheta_1 - \vartheta_2$, $\mu := m_2\ell_1\ell_2$, and $g = 9.81$. Truth: $(m_1, m_2, \ell_1, \ell_2) = (1, 0.5, 1, 0.8)$, $\sigma_r = 0.02$. Linearity in masses yields the invariant direction $(m_1, m_2) \to \alpha(m_1, m_2)$. Both angles are observed $K = 20$ times over $\mathcal{T} = 12$ ($\sigma_{\mathrm{obs}} = 0.02$\,rad). The conservative divergence integrates to a bounded boundary term~\eqref{eq:Model_Difference_OnsagerMachlup}, but chaos yields a $\Theta(\mathcal{T})$ surrogate determinant term (Appendix~\ref{sec:Exp-Conservative}).

\textbf{Relativistic particle} ($d=1$): rest mass $m_0$, charge $q$, accelerated by $E(t) = 1.81\cos(\pi t)$ up to $0.5c$. Drift: $m_0\gamma^3\ddot{x} = qE(t)$ with Lorentz factor $\gamma = (1 - \dot{x}^2/c^2)^{-1/2}$; truth: $(m_0, q, c) = (1, 1, 1)$, $\sigma_r = 0.02$. As only $q/m_0$ governs the drift, the invariant direction is $(m_0, q) \to \alpha(m_0, q)$. Position is observed $K = 20$ times over $\mathcal{T} = 12$ ($\sigma_{\mathrm{obs}} = 0.005$). The divergence integrates to a bounded boundary term but alternates sign, so $\sum_k |\mathrm{Re}\,s_k| = |\nabla\!\cdot\! f_\lambda|$ grows with $\mathcal{T}$ (Appendix~\ref{sec:Exp-Conservative}).

\paragraph{Evaluation grid.}
We evaluate two residual metrics—\emph{acceleration} ($A_\lambda = I_d$, metric bias off) and \emph{force/torque} ($A_\lambda = M_\lambda$, metric bias on)—across four inference schemes:
\textbf{Chain (state space)}: our method (Section~\ref{sec:Method}; PMMH, $1{,}000$--$2{,}000$ particles, $24{,}000$ iterations).
\textbf{Collider (state space)}: differs from Chain solely by omitting the normalizer.
\textbf{Collider B-PINN}: standard model (Section~\ref{sec:Background-Models}) with a Fourier-feature MLP ($\dim\theta \approx 1.4$k, $Nd \le 720$), jointly sampled over $(\theta, \lambda)$ via NUTS ($4$ chains, $9{,}000$ steps).
\textbf{Chain (Fourier MLP)}: augments the B-PINN with the closed-form correction $-\Delta$~\eqref{eq:TwoBiasesDecomposition} to restore $Z_\lambda$. We run this only on the oscillator, where state-independent $A_\lambda$ and $\nabla\!\cdot\! f_\lambda$ allow closed-form evaluation at each NUTS step; in the other systems, state dependence makes it prohibitive (Appendix~\ref{sec:BilionisComparison}).
Details and code: Appendix~\ref{sec:Exp-Details} and \url{https://github.com/obermm/ColliderBias}.


\begin{figure}[t]
    \centering
    \includegraphics[width=\columnwidth]{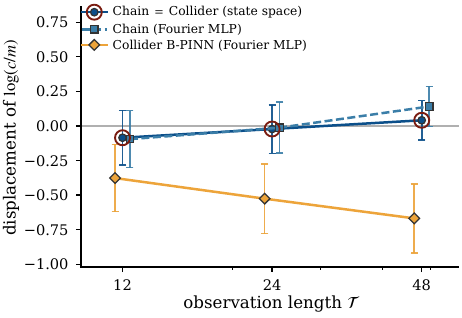}
    \caption{\textbf{Longer records amplify the dissipation bias} (oscillator, acceleration residuals). 
    Posterior mean of $\log(c/m)$ minus its true value, with $\pm 1$ posterior standard deviation. 
    The chain stays near the truth on both latents, and its posterior contracts as $\mathcal{T}$ grows; on the state-space latent the collider coincides with it (open rings). The B-PINN drifts further from the truth while its posterior width stays constant, so its error grows from $1.55$ to $2.67$ standard deviations.}
    \label{fig:Exp_Osc_Horizon}
\end{figure}

 
\subsection{Collider Bias and Its Removal} \label{sec:Exp-Grid}

In Table~\ref{tab:grid}, the chain model recovers the truth throughout: $|\log\alpha| \le 0.18$ ($\le 1.1$ posterior SDs) and identified parameters within $1.3$ SDs (Appendix~\ref{sec:Exp-FullGrid}). The collider fails as predicted.

\paragraph{Null condition.}
With acceleration residuals ($A_\lambda = I_d$) on the Markovian state-space latent, both biases are inactive: $\Delta(\lambda)$ in Eq.~\eqref{eq:Metric-Tilt-Exact} is constant, yielding bit-identical chain and collider draws.

\paragraph{Mechanism 1: residual metric bias.}
Under force residuals, the metric term in Eq.~\eqref{eq:TwoBiasesDecomposition} rewards small masses by adding $-Td\log\alpha$ to the state-space log-density and $\approx -n_{\mathrm{eff}}\log\alpha$ to the B-PINN, opposed along the invariant ray only by process noise (Appendix~\ref{sec:Ray-Collapse}).
On state-space latents ($T = 1{,}248$ for the oscillator), this reward dominates: the Kalman posterior collapses to $\log\alpha = -42.4$, and PMMH drifts toward zero mass until Euler integration fails. 
Measured $\Delta(\lambda)$ values match Eq.~\eqref{eq:Metric-Tilt-Exact} to machine precision (Appendix~\ref{sec:Exp-LawDetail}). 
On the B-PINN, a weaker reward ($n_{\mathrm{eff}} \approx 110$) is checked by noise, yet mass scales are still underestimated by factors of $2.5$ to $16$.
The damage is not confined to the drift-invariant direction: since $\det M_\lambda \propto \ell_1^2\ell_2^2$ also rewards short rods, the pendulum B-PINN shrinks $\ell_1$ and $\ell_2$ to $0.14$ and $0.27$ of their true values, so removing the scale ambiguity by reparameterization would not remove the bias.

\paragraph{Mechanism 2: dissipation bias.}
Omitting divergence~\eqref{eq:NonlocalBias} biases B-PINNs toward underestimating dissipation. On the oscillator with acceleration residuals, B-PINN underestimates damping ($c/m = 0.73$ vs.\ true $1.20$; Fig.~\ref{fig:Exp_Osc_Posteriors}).
This is systematic: across ten dataset resamples, B-PINN underestimates $\log(c/m)$ by $0.42 \pm 0.03$ and lies below the truth on every dataset, with $90\%$ credible intervals that cover the truth in only $4$ of $10$ cases, whereas our chain's intervals always do (Appendix~\ref{sec:Exp-Replication}).
With $\sigma_r$ matched to the data, the bias persists for four tested $\sigma_r \in [0.05, 1]$ (Appendix~\ref{sec:Exp-SigmaSweep}).
For the relativistic particle, B-PINN overestimates the speed of light ($c = 1.78$ vs.\ $1.00$): although the divergence integral vanishes, its absolute path integral does not, penalizing smaller $c$ by $9.5$\,nats per e-fold (Appendix~\ref{sec:Exp-Conservative}).
The pendulum is unassessed, as its B-PINN does not converge.

\paragraph{Longer records amplify dissipation bias.}
As the penalty scales as $\Theta(\mathcal{T})$, longer horizons ($\mathcal{T} = 12 \to 48$) push B-PINN damping estimates further from the truth without posterior contraction (Fig.~\ref{fig:Exp_Osc_Horizon}). 
Reweighting the chain posterior by $e^{\Delta_{\mathrm{diss}}}$ predicts the bias's direction, growth, and 60–75\% of its size (Appendix~\ref{sec:Exp-LawDetail}).

\paragraph{Restoring the normalizer on the surrogate.}
Chain (Fourier MLP) adds $-\Delta$ to an otherwise identical B-PINN. 
Under acceleration residuals on the oscillator, it removes $97.5\%$ of the $\log(c/m)$ bias. Under force residuals, where metric correction uses approximate $n_{\mathrm{eff}}$, it eliminates two-thirds of the mass bias (overshooting slightly; Appendix~\ref{sec:Exp-NetChain}). Thus, the missing normalizer, not the neural surrogate, causes the pathology.

\paragraph{Comparison with nested inference.}
PMMH is up to $13\times$ faster than the B-PINN, depending on the system state size (Appendix~\ref{sec:Exp-Setup}).
We further benchmark PMMH against unmodified public reimplementations of NSVI~\citep{Bilionis2} and NPSGLD~\citep{Bilionis3} on Euler--Maruyama latents with closed-form $Z_\lambda$ (Appendix~\ref{sec:BilionisComparison}). 
PMMH converges across all settings ($0.51$--$6.71$\,s/ESS). NPSGLD fails to converge, requiring $21\times$--$2{,}900\times$ more compute per ESS while underestimating posterior SDs by up to $30\times$. 
NSVI converges to a biased damping ratio on the oscillator ($c/m \approx 0.80$ vs.\ $1.20$) and raising its compute by a factor of 100 does not remove this bias.
 
 
\section{Conclusions} \label{sec:Conclusions}
B-PINNs suffer from an overlooked structural defect: conditioning on virtual residual observations turns physical parameters $\lambda$ and trajectory surrogates into colliders. 
This implicitly neglects the parameter-dependent normalizing constant $Z_\lambda$, systematically biasing posterior inference. 
Even when initialized at the ground truth, standard B-PINNs underestimate damping and mass, and grow increasingly overconfident with longer observation records.

We approximated two bias mechanisms in closed form, confirmed them across three systems, yet showed that analytical corrections can remain insufficient. 
To eliminate collider bias entirely, we reformulate the trajectory as a Markovian state-space model. This enables an asymptotically exact particle MCMC scheme that avoids both the computational overhead of nested Langevin dynamics~\citep{Bilionis3} and the variational gap of nested VI~\citep{Bilionis2}.

\paragraph{When are B-PINNs safe?}
Collider B-PINNs yield unbiased posteriors if two null conditions hold simultaneously (Section~\ref{sec:Theory}; Appendix Table~\ref{tab:bias-taxonomy}): 
(i) the residual metric $A_\lambda$ is parameter-independent (e.g., measuring acceleration rather than force); and 
(ii) the dissipation bias~\eqref{eq:NonlocalBias} is negligible, with $\Delta_{\mathrm{diss}}$ remaining roughly constant across posterior support. 
Otherwise, practitioners must adopt a Markovian chain formulation or, when $A_\lambda$ and $\nabla\!\cdot\! f_\lambda$ are state-independent, apply the closed-form correction $-\Delta$ directly.

\paragraph{Limitations and Future Work.}
Our particle MCMC scheme targets only stochastic ODEs ($\Sigma_r \succ 0$) in low-to-moderate dimensions; particle filters degenerate as $\sigma_r \to 0$ and face weight collapse on finely discretized PDEs. 
Because collider bias persists in PDE systems, extending our framework to spatial domains will require scalable alternatives such as ensemble Kalman or score-based filtering. 
Furthermore, selecting $\Sigma_r$ to prevent residual misspecification remains open. 
Ideally $\Sigma_r$ would be inferred jointly with $\lambda$, but this introduces a third collider-bias mechanism that we leave to future work.

\paragraph{Implications.}
Although standard B-PINNs rely on collider structures, the resulting bias has gone largely unnoticed, as they appear well-behaved when the null conditions approximately hold. 
When violated, MCMC still converges cleanly to confidently wrong posteriors whose error is undetectable without ground truth. 
Crucially, this distortion peaks in weakly identified regimes and long horizons with sparse data—precisely when Bayesian rigor matters most.
Practitioners should therefore check these null conditions and prefer a chain formulation.
 
\section*{Acknowledgements}
This research was funded by the Austrian Research Promotion Agency (FFG) via the project VENTUS (910263) under the FFG AI for Green program.
We thank Ilias Bilionis, Atharva Hans, and Maxwell Stephen Bolt for kindly sharing their implementation of the NSVI and NPSGLD algorithms used as benchmarks in this work.

\clearpage
 
\bibliographystyle{plainnat} 
\bibliography{FixPINNs}
 
 
\clearpage
\appendix
 
\thispagestyle{empty}

\onecolumn
\section{SUPPLEMENTARY MATERIAL}
 
 
\subsection{Laplace Approximation of the Collider Bias for Global Surrogates} \label{sec:Tilt-Derivation}
This appendix derives the two-term decomposition~\eqref{eq:TwoBiasesDecomposition} for a global surrogate such as a neural network.
Step~1 approximates $Z_\lambda$ by Laplace's method, Step~2 identifies the two factors through which $\lambda$ enters the resulting curvature, Steps~3 and~4 differentiate with respect to these factors, and Step~5 collects the terms; a worked example for the oscillator follows.
Along drift-invariant directions, the metric bias admits exact statements, derived in Appendix~\ref{sec:Ray-Collapse}.

\paragraph{Setup.}
We write the residual~\eqref{eq:Residual-Unified} in continuous time,
\begin{equation} \label{eq:Tilt-Residual}
    r_\theta(t,\lambda) := A_\lambda(z_\theta(t)) \left( \dot{v}_\theta(t) - a_\lambda(z_\theta(t), t) \right),
    \qquad v_\theta := \dot{x}_\theta,
\end{equation}
so that $r_n(\theta,\lambda) = r_\theta(t_n,\lambda)$; since $v_\theta$ is the derivative of $x_\theta$, only the acceleration equation can be violated.
The surrogate carries the weight prior $p(\theta) = \mathcal{N}(0, P^{-1})$ of Section~\ref{sec:Theory}, where $P \to 0$ corresponds to a flat reference measure.
Chain and collider share this prior and the energy~\eqref{eq:EnergyFunctional} and differ only in their normalizers, $Z_\lambda$ in~\eqref{eq:Z-Integral} and $Z_r$ in~\eqref{eq:BPINN-Likelihood}, so that $\Delta = \log Z_\lambda - \log Z_r + \mathrm{const}$ by~\eqref{eq:Delta-Definition}.
Only $Z_\lambda$ depends on $\lambda$; since $\Sigma_r$ is fixed, $\log Z_r$ is a constant.

\paragraph{Step 1: Laplace approximation.}
Let $\theta^*(\lambda)$ minimize $E[\theta, \lambda] - \log p(\theta)$, the negative logarithm of the data-free integrand of $Z_\lambda$.
A second-order expansion around $\theta^*$ gives
\begin{equation} \label{eq:Tilt-Laplace}
    \log Z_\lambda \approx -E[\theta^*, \lambda] - \frac{1}{2}\theta^{*\top} P \theta^* - \frac{1}{2}\log\det H_\lambda + \mathrm{const}, \qquad H_\lambda := \nabla^2_\theta E\big|_{\theta^*} + P ,
\end{equation}
where the constant ($\tfrac12 \log\det P$) does not depend on $\lambda$.
We drop the first two terms, assuming that the surrogate (nearly) solves the ODE, $E[\theta^*, \lambda] \approx 0$, and that $\theta^*$ follows an ODE solution rather than the prior mode, so that $\tfrac12\theta^{*\top} P \theta^*$ depends only weakly on $\lambda$.
This is exact for linear, unforced dynamics and a surrogate with $x_\theta \equiv 0$ at $\theta = 0$, where $\theta^* = 0$ for every $\lambda$.
Since the residuals vanish at $\theta^*$, so does the part of $\nabla^2_\theta E$ that is proportional to them, which leaves $H_\lambda \approx G_\lambda + P$ with the Gauss--Newton curvature
\begin{equation} \label{eq:Tilt-GaussNewton}
    G_\lambda := h \sum_{n=1}^N \left(\frac{\partial r_\theta(t_n,\lambda)}{\partial\theta}\right)^{\!\top} \Sigma_r^{-1}\, \frac{\partial r_\theta(t_n,\lambda)}{\partial\theta},
    \qquad
    \log Z_\lambda \approx -\frac{1}{2}\log\det(G_\lambda + P) + \mathrm{const}.
\end{equation}
Since $\log Z_r$ is constant, this gives Eq.~\eqref{eq:Delta-Expansion}; the remaining steps evaluate its first term.

\paragraph{Step 2: Two factors of the curvature.}
Let $B(t) := \partial z_\theta(t)/\partial\theta$ be the tangent map of the surrogate at $\theta^*$, and $Df_\lambda(t) := \partial f_\lambda/\partial z$ the drift Jacobian along the fitted trajectory $z(t) := z_{\theta^*}(t)$.
By the chain rule,
\begin{equation} \label{eq:Tilt-Jacobian}
    \frac{\partial r_\theta(t,\lambda)}{\partial\theta} \approx A_\lambda(z(t))\, (L_\lambda B)_v(t),
    \qquad
    L_\lambda := \frac{\mathrm{d}}{\mathrm{d}t} - Df_\lambda(t),
\end{equation}
where $(\cdot)_v$ denotes the velocity block.
The position block of $L_\lambda B$ vanishes because $v_\theta = \dot{x}_\theta$, and the term containing the derivative of $A_\lambda$ multiplies the residual, which vanishes at $\theta^*$.
Stacking over the collocation points gives
\begin{equation} \label{eq:Tilt-Factorization}
    G_\lambda = \mathbf{B}^\top \mathbf{L}_\lambda^\top \mathbf{A}_\lambda^\top\, \mathbf{S}^{-1}\, \mathbf{A}_\lambda \mathbf{L}_\lambda \mathbf{B},
    \qquad
    \mathbf{A}_\lambda = \operatorname{diag}\big(A_\lambda(z(t_1)), \dots, A_\lambda(z(t_N))\big),
    \quad
    \mathbf{S} = h^{-1}\operatorname{diag}(\Sigma_r, \dots, \Sigma_r).
\end{equation}
The stacked tangent map $\mathbf{B}$ depends on $\lambda$ only through $\theta^*(\lambda)$, which we hold fixed.
Since $\mathbf{S}$ is fixed, $\lambda$ enters $G_\lambda$ through two factors: the residual metric $\mathbf{A}_\lambda$ and the drift operator $\mathbf{L}_\lambda$.
Since these factors are rectangular, $\log\det(G_\lambda + P)$ does not split into a sum over them; we therefore differentiate it with respect to each factor in turn, holding the other fixed.

\paragraph{Step 3: Residual metric (Mechanism~1).}
Let $A_\lambda(z) = \kappa_\lambda \bar{A}(z)$ with a scalar $\kappa_\lambda > 0$ and a $\lambda$-independent $\bar{A}$, e.g., $A_\lambda = m$ for the oscillator.
Then the metric scale factors out of~\eqref{eq:Tilt-Factorization},
\begin{equation} \label{eq:Tilt-u}
    G_\lambda = \kappa_\lambda^2\, \tilde{G},
    \qquad
    \tilde{G} := \mathbf{B}^\top \mathbf{L}_\lambda^\top \bar{\mathbf{A}}^\top \mathbf{S}^{-1} \bar{\mathbf{A}} \mathbf{L}_\lambda \mathbf{B},
\end{equation}
where $\bar{\mathbf{A}} := \operatorname{diag}\big(\bar{A}(z(t_1)), \dots, \bar{A}(z(t_N))\big)$, and $\tilde{G}$ does not depend on $\kappa_\lambda$.
Since $\partial G_\lambda/\partial\log \kappa_\lambda = 2 G_\lambda$, Jacobi's formula gives
\begin{equation} \label{eq:Tilt-VolumeDerivative}
    \frac{\partial}{\partial\log \kappa_\lambda}\left(-\frac{1}{2}\log\det(G_\lambda + P)\right)
    = -\operatorname{tr}\!\left[(G_\lambda + P)^{-1} G_\lambda\right]
    = -\,n_{\mathrm{eff}}.
\end{equation}

\emph{When is $n_{\mathrm{eff}}$ constant?}
For $P \succ 0$, let $\{g_k\}$ be the eigenvalues of $P^{-1/2}\tilde{G}P^{-1/2} \succeq 0$.
Then $\det(G_\lambda + P) = \det(P)\prod_k(1 + g_k \kappa_\lambda^2)$, so that
\begin{equation} \label{eq:Tilt-Neff-Spectrum}
    -\frac{1}{2}\log\det(G_\lambda + P) = -\frac{1}{2}\sum_k \log\!\left(1 + g_k \kappa_\lambda^2\right) + \mathrm{const},
    \qquad
    n_{\mathrm{eff}} = \sum_k \frac{g_k \kappa_\lambda^2}{1 + g_k \kappa_\lambda^2}.
\end{equation}
Each summand lies in $[0,1)$ and is nonzero only for $g_k > 0$, so $n_{\mathrm{eff}} < \operatorname{rank}G_\lambda \le \min(Nd, \dim\theta)$ whenever $G_\lambda \neq 0$.
Once the physics dominates the prior on every direction it constrains ($g_k \kappa_\lambda^2 \gg 1$), $n_{\mathrm{eff}}$ saturates at $\operatorname{rank}G_\lambda$ and no longer depends on $\kappa_\lambda$; we call this the saturated regime.
For $P = 0$ and full-rank $G_\lambda$ (which requires $\dim\theta \le Nd$), $n_{\mathrm{eff}} = \dim\theta$ for every $\kappa_\lambda$.

\emph{Mechanism~1 (residual metric).}
Since $Z_r$ does not depend on $\kappa_\lambda$, Eq.~\eqref{eq:Tilt-VolumeDerivative} gives $\partial\Delta/\partial\log\kappa_\lambda = -n_{\mathrm{eff}}$.
With $\sum_n \log|\det A_\lambda(z(t_n))| = Nd\log\kappa_\lambda + \mathrm{const}$, this integrates, for constant $n_{\mathrm{eff}}$, to the metric term $-\frac{n_{\mathrm{eff}}}{Nd}\sum_{n}\log|\det A_\lambda(z(t_n))|$ of~\eqref{eq:TwoBiasesDecomposition}.
For a state-dependent $A_\lambda(z(t))$, the term holds locally: on time windows that are short compared with the variation of $A_\lambda(z(t))$ but long compared with the resolution of the surrogate, $A_\lambda$ is nearly constant, and if the constrained directions are spread uniformly over time and components, each window contributes its share $n_{\mathrm{eff}}/Nd$ of the local sum of $\log|\det A_\lambda|$.
This neglects the coupling between windows and the anisotropy of $A_\lambda$ relative to $\Sigma_r$, both absent in the scalar case.

\emph{State-space latent.}
On the Markovian latent of Section~\ref{sec:Method}, $\theta$ consists of the $Td$ velocities under a flat prior ($P = 0$), and $\partial r/\partial v$ is block lower-triangular with invertible diagonal blocks $A_\lambda(z_{i-1})/h$.
Hence $n_{\mathrm{eff}} = Td = Nd$ exactly, and the metric term becomes the exact collider bias~\eqref{eq:Metric-Tilt-Exact}.

\paragraph{Step 4: Drift operator (Mechanism~2).}
We set $A_\lambda = I_d$, so that $\lambda$ enters~\eqref{eq:Tilt-Factorization} only through $\mathbf{L}_\lambda$.
We assume that the span of the surrogate's tangent map is approximately invariant under $L_\lambda$, with a $\lambda$-independent Gram determinant, and that the physics dominates $P$ on the resolved modes.
Then $\frac{1}{2}\log\det(G_\lambda + P)$ equals $\log|\det L_\lambda|$ restricted to the resolved modes, up to $\lambda$-independent terms.
On the position component, $L_\lambda$ acts as $\frac{\mathrm{d}^2}{\mathrm{d}t^2} - \frac{\partial a_\lambda}{\partial v}\frac{\mathrm{d}}{\mathrm{d}t} - \frac{\partial a_\lambda}{\partial x}$, whose matrix symbol at frequency $\omega$ has, by the Schur complement, the same determinant as $\hat{L}_\lambda(\omega) := i\omega I_{2d} - Df_\lambda$.
For constant $Df_\lambda$ (relaxed below), a surrogate with effective bandwidth $\Omega$ on $[0,\mathcal{T}]$ resolves real Fourier modes at density $\mathcal{T}/\pi$, so that
\begin{equation} \label{eq:Tilt-Symbol}
    \frac{1}{2}\log\det(G_\lambda + P) \;\approx\; \frac{\mathcal{T}}{\pi}\int_0^{\Omega} \log\left|\det \hat{L}_\lambda(\omega)\right| \mathrm{d}\omega .
\end{equation}

\emph{The eigenvalue integral.}
Let $\{s_k\}_{k=1}^{2d}$ be the eigenvalues of $Df_\lambda$, so that $\log|\det\hat L_\lambda(\omega)| = \sum_k \log|i\omega - s_k|$.
The integral~\eqref{eq:Tilt-Symbol} diverges as $\Omega\to\infty$, but at a rate independent of $\lambda$; differentiating with respect to the eigenvalues isolates the $\lambda$-dependent part.
For a complex pair $s = -\rho \pm i\beta$ with $\beta > 0$ ($\rho > 0$ damped, $\rho < 0$ growing), $I(\rho,\beta) := \int_0^\Omega \log|(i\omega-s)(i\omega-\bar s)|\,\mathrm{d}\omega$ satisfies
\begin{align}
    \frac{\partial I}{\partial \rho}
    &= \int_0^\Omega \left( \frac{\rho}{(\omega-\beta)^2+\rho^2} + \frac{\rho}{(\omega+\beta)^2+\rho^2} \right) \mathrm{d}\omega
    = \arctan\tfrac{\Omega-\beta}{\rho} + \arctan\tfrac{\Omega+\beta}{\rho}
    \;=\; \pi\,\operatorname{sign}(\rho) + \mathcal{O}(1/\Omega), \label{eq:Tilt-dIdalpha} \\
    \frac{\partial I}{\partial \beta} &= \frac{1}{2}\log\frac{(\Omega+\beta)^2+\rho^2}{(\Omega-\beta)^2+\rho^2} = \mathcal{O}(\beta/\Omega), \label{eq:Tilt-dIdbeta}
\end{align}
where the boundary terms at $\omega = 0$ cancel.
Up to $\mathcal{O}(1/\Omega)$, the damping rate thus enters with weight $\pi$ and the frequency $\beta$ not at all, so each pair contributes $\pi|\rho|$.
A real eigenvalue $s = -\rho$ gives $\partial_\rho \int_0^\Omega \log|i\omega + \rho|\,\mathrm{d}\omega = \arctan(\Omega/\rho) = \frac{\pi}{2}\operatorname{sign}(\rho) + \mathcal{O}(1/\Omega)$, i.e., a contribution $\frac{\pi}{2}|\rho|$.
Every eigenvalue therefore contributes $\frac{\pi}{2}|\mathrm{Re}\, s_k|$, and multiplying by the mode density $\mathcal{T}/\pi$ gives
\begin{equation} \label{eq:Tilt-Diss-General}
    \frac{1}{2}\log\det(G_\lambda + P) = \frac{\mathcal{T}}{2}\sum_{k=1}^{2d} |\mathrm{Re}\, s_k| + \mathrm{const} + \mathcal{R},
    \qquad \mathcal{R} = \mathcal{O}(\mathcal{T}/\Omega).
\end{equation}
Without growing modes ($\mathrm{Re}\, s_k \le 0$ for all $k$), $\sum_k |\mathrm{Re}\, s_k| = -\operatorname{tr}(Df_\lambda) = -\nabla\!\cdot\! f_\lambda$, and inserting~\eqref{eq:Tilt-Diss-General} into~\eqref{eq:Delta-Expansion} gives
\begin{equation} \label{eq:Tilt-Diss-Result}
    \Delta_{\mathrm{diss}}(\lambda) = \frac{\mathcal{T}}{2}\,\nabla\!\cdot\! f_\lambda + \mathrm{const} + \mathcal{R} .
\end{equation}
Otherwise, $\Delta(\lambda)$ is not determined by the divergence alone: growing modes are penalized like damped ones, and a divergence that changes sign along the orbit enters through its absolute value.
Appendix~\ref{sec:Exp-Conservative} discusses both cases for conservative systems.

\emph{Nonlinear drift.}
For nonlinear dynamics, we treat $Df_\lambda(z(t))$ as constant on windows that are short compared with $\mathcal{T}$ but long compared with the correlation time $\Omega^{-1}$ of the surrogate and the relaxation times $|\mathrm{Re}\,s_k|^{-1}$.
Summing over windows replaces $\mathcal{T}\,\nabla\!\cdot\! f_\lambda$ by $\int_0^{\mathcal{T}} \nabla\!\cdot\! f_\lambda(z(t),t)\,\mathrm{d}t$ and gives Eq.~\eqref{eq:NonlocalBias}.
Its leading term is the Onsager--Machlup correction to the path action~\citep{OnsagerMachlup1953, Durr1978_OM}, obtained here as the fluctuation determinant of a band-limited surrogate.
The remainder $\mathcal{R}$ collects the $\mathcal{O}(1/\Omega)$ tails of~\eqref{eq:Tilt-dIdalpha} and~\eqref{eq:Tilt-dIdbeta}, the residual floor $E[\theta^*,\lambda]$ of a finite-capacity fit, and the windowing.
Since all of these grow with $\mathcal{T}$ at the same rate as the leading term, their relative size is set by $\Omega$ compared with the time scales of the system, not by the length of the record.

\emph{State-space latent.}
Under the Euler--Maruyama transition~\eqref{eq:EM}, $v_i$ depends only on the noise increments $\Delta W_j := \sqrt{h}\,\xi_j$ up to step $i$, so the Jacobian of $(\Delta W_1, \dots, \Delta W_T) \mapsto (v_1, \dots, v_T)$ is block lower-triangular with diagonal blocks $A_\lambda(z_{i-1})^{-1}\Sigma_r^{1/2}$, which do not involve the drift.
This is why the exact collider bias~\eqref{eq:Metric-Tilt-Exact} has no dissipation term: the Markovian latent normalizes the path density step by step (the It\^o pre-point form), whereas a global surrogate, which imposes no ordering, yields the midpoint (Onsager--Machlup) form, consistent with the Wong--Zakai limit of smooth approximations to Brownian paths~\citep{WongZakai1965}.

\paragraph{Step 5: Collecting the terms.}
Adding the metric term of Step~3 and the dissipation term of Step~4 gives Eq.~\eqref{eq:TwoBiasesDecomposition}; without growing modes, its dissipation term is the divergence integral~\eqref{eq:NonlocalBias}.
Of the four simplifications listed in Section~\ref{sec:Theory}, three need a caveat in practice.
Simplification~(i) fails when the data-free integrand of $Z_\lambda$ is dominated by the prior rather than the physics; the terms dropped from~\eqref{eq:Tilt-Laplace} then depend on $\lambda$.
For~(ii), when $x \equiv 0$ solves the ODE, $\theta^* = 0$, where the tangent map of a neural surrogate degenerates; we therefore evaluate the curvature at a representative state on the solution manifold (Appendix~\ref{sec:Exp-NetChain}).
Since the factors in~\eqref{eq:Tilt-Factorization} are rectangular, the additivity~(iv) holds only to first order.
Eq.~\eqref{eq:TwoBiasesDecomposition} thus captures the dominant dependence of $\Delta$ on $\lambda$ and shows where the collider bias is large, but adding $-\Delta$ to the log-posterior of a B-PINN need not recover the chain posterior exactly.
Table~\ref{tab:bias-taxonomy} summarizes both mechanisms.

\begin{table}[h]
    \caption{Taxonomy of structural biases in collider B-PINNs. 
    Scaling denotes the asymptotic rate of the collider bias $\Delta = \log(p_{\mathrm{collider}}/p_{\mathrm{chain}})$. 
    Null conditions indicate when $\Delta$ is invariant to inferred parameters; $\{s_k\}$ are eigenvalues of $Df_\lambda$.}
    \label{tab:bias-taxonomy}
    \vskip 0.05in
    \begin{center}
    \begin{small}
    \begin{tabularx}{0.85\textwidth}{@{} >{\raggedright\arraybackslash}p{2.0cm} >{\raggedright\arraybackslash}X >{\raggedright\arraybackslash}p{1.6cm} >{\raggedright\arraybackslash}X @{}}
        \toprule
        \textbf{Mechanism} & \textbf{Physical Origin} & \textbf{Scaling} & \textbf{Null Condition} \\
        \midrule
        \textbf{Residual Metric} & State-space map $A_\lambda$ depends on $\lambda$ & $\Theta(n_{\mathrm{eff}})$ or $\Theta(\mathcal{T})$ & $A_\lambda$ invariant to $\lambda$ \\
        \addlinespace
        \textbf{Dissipation} & Phase-space volume contraction & $\Theta(\mathcal{T})$ & Markovian latent, or $\sum_k \lvert\operatorname{Re} s_k\rvert$ invariant to $\lambda$ \\
        \bottomrule
    \end{tabularx}
    \end{small}
    \end{center}
    \vskip -0.1in
\end{table}

\paragraph{Worked example: the damped linear oscillator.}
For $m\ddot{x}+c\dot{x}+kx=\sigma_r\xi$ with force residuals ($A_\lambda = m$, $d = 1$), the drift $f_\lambda(x,v) = (v,\, -\tfrac{k}{m}x-\tfrac{c}{m}v)^\top$ has the constant Jacobian $Df_\lambda = \begin{bsmallmatrix} 0 & 1 \\ -k/m & -c/m \end{bsmallmatrix}$ with divergence $\nabla\!\cdot\! f_\lambda = -c/m$.
Its eigenvalues are $-\rho \pm i\beta$ with $\rho = c/(2m)$ and $\beta = \sqrt{k/m - c^2/(4m^2)}$ if underdamped, and real and negative if overdamped; in both cases $\sum_k|\mathrm{Re}\,s_k| = c/m$.
No windowing is needed, and~\eqref{eq:TwoBiasesDecomposition} reads
\begin{equation} \label{eq:Tilt-Osc-Delta}
    \Delta(\lambda) \approx
    \underbrace{-\,n_{\mathrm{eff}}\log m}_{\text{metric}}
    \; \underbrace{-\; \frac{c\,\mathcal{T}}{2m}}_{\text{dissipation}}
    \; + \; \mathrm{const} .
\end{equation}
The dissipation term penalizes large $c/m$ in proportion to $\mathcal{T}$, so the collider underestimates damping.
The stiffness enters only through $\beta$, which by~\eqref{eq:Tilt-dIdbeta} contributes no leading-order term; the $\mathcal{O}(\beta/\Omega)$ remainder, however, still grows with $\mathcal{T}$ and rewards smaller $k/m$, so restoring only the leading-order terms leaves part of the stiffness bias (Appendix~\ref{sec:Exp-NetChain}).
Along the drift-invariant rescaling $(m,c,k)\to\alpha(m,c,k)$, only the metric term varies, $\partial\Delta/\partial\log\alpha = -n_{\mathrm{eff}}$: the collider is rewarded without bound for shrinking the mass scale, opposed only by the prior and by the process-noise level, which identifies $\sigma_r/m$.
On the state-space latent, the same holds exactly with $n_{\mathrm{eff}} = N = T$, since~\eqref{eq:Metric-Tilt-Exact} gives $\Delta_{\mathrm{Markov}} = -T\log m + \mathrm{const}$; Appendix~\ref{sec:Ray-Collapse} discusses its effect on the collider posterior.


\subsection{Exact Collider Bias along Drift-Invariant Directions} \label{sec:Ray-Collapse}
While the decomposition derived in Appendix~\ref{sec:Tilt-Derivation} is generally approximate, the metric bias (Mechanism~1) admits exact statements along drift-invariant directions. 

Consider a drift-invariant direction parameterized by jointly scaling masses and forces by a factor $\alpha$.
Suppose the scaled components follow independent log-normal priors with log-scale standard deviations $s_j$. 
Measuring $\log\alpha$ relative to the prior medians $\tilde\lambda_j$ via the precision-weighted mean
\begin{equation} \label{eq:Ray-alpha}
    \log\alpha = \frac{\sum_j s_j^{-2} \log(\lambda_j/\tilde\lambda_j)}{\sum_j s_j^{-2}},
\end{equation}
the prior factorizes into independent components along $\log\alpha$ and the identified directions, with a Gaussian marginal of standard deviation $(\sum_j s_j^{-2})^{-1/2}$ along $\log\alpha$.

Under acceleration residuals, the residuals themselves remain invariant under this scaling.
Consequently, neither the residual energy~\eqref{eq:EnergyFunctional} nor the observational likelihood depends on $\alpha$, and $Z_\lambda$ is independent of $\alpha$. 
Both the chain and the collider joint then depend on $\alpha$ solely through the prior, so both posteriors leave the distribution of $\log\alpha$ exactly at its prior, for any latent representation.

Under force residuals, $A_\lambda = M_\lambda$ scales with $\alpha$ while $\Sigma_r$ is fixed, so the process noise on the velocities scales with $\alpha^{-1}$: the data constrain $\alpha$ through the noise level, and the chain posterior of $\log\alpha$ differs from its prior. 
On top of this, the collider joint acquires an additional explicit $\alpha$-dependence: on the state-space latent, $\Delta_{\mathrm{Markov}} = -Td\log\alpha + \mathrm{const}$ exactly by Eq.~\eqref{eq:Metric-Tilt-Exact}, and on a global surrogate, $\Delta \approx -n_{\mathrm{eff}}\log\alpha + \mathrm{const}$ by the metric term of Eq.~\eqref{eq:TwoBiasesDecomposition}.
The collider posterior of $\log\alpha$ is therefore the chain posterior reweighted by $\alpha^{-Td}$ (respectively $\alpha^{-n_{\mathrm{eff}}}$).
This reward grows with the number of time steps (respectively constrained directions), whereas the constraint from the noise level is set by the finitely many observations; on the oscillator, where the exact Kalman likelihood is available, it places the state-space collider at $\log\alpha = -42.4$ (Table~\ref{tab:grid}).

 
\subsection{Experimental Details and Full Results} \label{sec:Exp-Details}

This appendix supports Section~\ref{sec:Experiments} and follows its order.
Appendix~\ref{sec:Exp-Setup} specifies the setup, and Appendix~\ref{sec:Exp-Validation} verifies that our method samples the chain posterior correctly, since every bias in this paper is measured against it.
Appendix~\ref{sec:Exp-FullGrid} reports all parameters of the runs summarized in Table~\ref{tab:grid}, and Appendix~\ref{sec:Exp-Replication} shows that the bias on the oscillator is reproducible across datasets, and Appendix~\ref{sec:Exp-SigmaSweep} how it depends on the residual scale.
The next two subsections support the discussion of individual mechanisms in Section~\ref{sec:Exp-Grid}: the dissipation term on the two conservative systems (Appendix~\ref{sec:Exp-Conservative}) and the closed-form correction of the neural surrogate (Appendix~\ref{sec:Exp-NetChain}).
Appendix~\ref{sec:Exp-LawDetail} details the scaling-law experiments of Figs.~\ref{fig:Exp_Osc_Horizon} and~\ref{fig:Exp_Osc_Exact}.

\subsubsection{Setup} \label{sec:Exp-Setup}

\paragraph{Evaluation grid.}
Each of the three systems is combined with two residual metrics (acceleration and force/torque residuals) and with the inference variants listed below.
In every configuration, the process noise is isotropic, $\Sigma_r = \sigma_r^2 I_d$, with $\sigma_r$ fixed at its true value.
Within a system, all runs share the dataset, ground truth and prior, the time discretization, number of particles, sampler budget and random seed, and the network and collocation grid of the surrogate.
Two runs that differ in a single factor of the grid therefore isolate the effect of that factor, for instance of the residual metric.

\paragraph{Systems and data.}
Table~\ref{tab:system-settings} lists the settings that differ between systems.
The data are simulated with the same semi-implicit Euler--Maruyama scheme~\eqref{eq:EM} that the inference uses, but on a finer time grid: $2\times$ finer for the oscillator and about $50\times$ finer for the pendulum and the relativistic particle (Appendix~\ref{sec:Exp-Validation} shows that the results do not rely on this match).
The number of Euler--Maruyama steps $T$ used for inference and the number of particles were chosen per system by a convergence study, and are identical across all state-space runs of that system, so that chain and collider remain comparable.

\begin{table}[h]
\caption{\textbf{Per-system settings.} The MLP surrogate maps a Fourier-feature embedding of dimension $33$ through two hidden layers of width $24$ to the $d$ output coordinates. $Nd$ is the number of residual components, i.e., collocation points times dimension.}
\label{tab:system-settings}
\vskip 0.05in
\begin{center}
\begin{small}
\begin{tabular}{@{}lrrr@{}}
\toprule
 & \textbf{Oscillator} & \textbf{Double pendulum} & \textbf{Relativistic particle} \\
\midrule
ground truth $\lambda$ & $(m,c,k) = (1,\,1.2,\,4)$ & $(m_1,m_2,\ell_1,\ell_2) = (1,\,0.5,\,1,\,0.8)$ & $(m_0,q,c) = (1,\,1,\,1)$ \\
residual scale $\sigma_r$ (fixed) & $0.30$ & $0.02$ & $0.02$ \\
horizon $\mathcal{T}$ & $24$ & $12$ & $12$ \\
observations $K$ & $40$ & $20$ & $20$ \\
observation noise $\sigma_{\mathrm{obs}}$ & $0.02$ & $0.02$ rad & $0.005$ \\
Euler--Maruyama steps $T$ & $1248$ & $1235$ & $12160$ \\
particles & $1000$ & $1000$ & $2000$ \\
\midrule
MLP layer widths & 33-24-24-1 & 33-24-24-2 & 33-24-24-1 \\
$\dim\theta$ & $1441$ & $1466$ & $1441$ \\
residual components $Nd$ & $128$ & $720$ & $300$ \\
\bottomrule
\end{tabular}
\end{small}
\end{center}
\end{table}

\paragraph{Inference variants.}
\begin{itemize}
    \item \textbf{Chain (state space)} is our method (Section~\ref{sec:Method}). Its latent is the Euler--Maruyama path $z_{0:T}$ with the normalized transition kernel~\eqref{eq:Chain-TransitionKernel}; a bootstrap particle filter estimates the marginal likelihood $\hat p(y\mid\lambda)$, and PMMH samples $\lambda$.
    \item \textbf{Collider (state space)} uses the same latent, particle filter and sampler, but omits $Z_\lambda$. It is a control that differs from our method only in the normalizer.
    \item \textbf{Collider B-PINN} is the standard formulation of Section~\ref{sec:Background-Models}: a Fourier-feature MLP surrogate $x_\theta$ with a Gaussian weight prior, sampled with NUTS jointly over $(\theta, \lambda)$.
    \item \textbf{Chain (Fourier MLP)} is the Collider B-PINN with the closed-form correction $-\Delta$ for the missing normalizer added to its log-density (Appendix~\ref{sec:Exp-NetChain}). It is run on the oscillator only (Section~\ref{sec:Exp-Systems}).
\end{itemize}
Because the oscillator is linear, the marginal likelihood of its state-space latent is also available exactly from a Kalman filter.
We use this in one place.
Under force residuals, PMMH cannot reach the posterior of the state-space collider (see below), so on the oscillator we sample it with an adaptive random-walk Metropolis sampler instead, targeting the exact chain likelihood multiplied by $e^{\Delta(\lambda)} = m^{-T}$ from Eq.~\eqref{eq:Metric-Tilt-Exact} ($4$ chains, $\hat{R} \le 1.001$).
These entries are marked with $\ddagger$.
Applied to the chain model, the same sampler reproduces the PMMH posterior (e.g., $\log m$: $0.014 \pm 0.139$ against $0.006 \pm 0.138$).

\paragraph{Priors.}
All physical parameters have log-normal priors with median at the ground truth.
The standard deviations of the log-parameters are $0.4$ for masses, charges and the damping coefficient $c$ of the oscillator, and $0.25$ for lengths, stiffnesses and the light speed $c$ of the relativistic particle.
The Gaussian weight prior of the B-PINN follows LeCun scaling at unit gain: each weight of a layer with $n_{\mathrm{in}}$ inputs has variance $1/n_{\mathrm{in}}$, and each bias has variance $1$.

\paragraph{Samplers.}
PMMH runs for $24{,}000$ iterations.
The first $4{,}000$ adapt the proposal, and the first $30\%$ of the remaining $20{,}000$ are discarded as burn-in, which leaves $14{,}000$ draws.
NUTS runs $4$ chains, each with $3{,}000$ warm-up iterations and $6{,}000$ draws at a maximum tree depth of $11$, which gives $24{,}000$ draws in total.
All samplers use random seed $0$.
Each run uses a single GPU (NVIDIA Quadro RTX 8000 or NVIDIA L40) on an internal compute cluster.
On the oscillator, for instance, PMMH takes $442$\,s and the B-PINN $5{,}834$\,s (NUTS) for these budgets.

\paragraph{Software and licenses.}
All methods are implemented in JAX~\citep{jax2018github}; NUTS and the autocorrelation-based effective sample sizes use NumPyro~\citep{phan2019numpyro, bingham2019pyro}.
The nested baselines of Appendix~\ref{sec:BilionisComparison} use the public reimplementation of \citet{Bolt2026_NIFFReplication}, which additionally depends on Optax.
JAX, NumPyro and Optax are released under the Apache License~2.0; the reimplementation does not state a license.
We release our code under the MIT license.

\paragraph{Reporting conventions.}
Unless stated otherwise, tables report the posterior mean divided by the ground truth, so that $1.00$ means no error.
There are two exceptions.
Table~\ref{tab:grid} reports the drift-invariant direction as the precision-weighted log scale factor $\log\alpha$ of Eq.~\eqref{eq:Ray-alpha}.
Tables~\ref{tab:replication} and~\ref{tab:netchain} report \emph{displacements} in log space, i.e., the posterior mean of the log parameter minus the log of its true value.
Since the mean of a logarithm is not the logarithm of the mean, these displacements are not exactly the logarithms of the ratios reported elsewhere; for instance, the damping ratio of the Collider B-PINN under acceleration residuals has a displacement of $-0.526$ (Table~\ref{tab:netchain}), and $\exp(-0.526) = 0.59$, whereas Table~\ref{tab:grid} reports the ratio $0.61$.

\paragraph{Runs without valid posterior samples ($\dagger$).}
Two kinds of runs do not yield valid posterior samples.
The first is the state-space collider under force residuals.
Along the drift-invariant direction, its log-density gains $-Td\log\alpha$, which rewards small masses.
Together with the log-normal prior, this still defines a proper posterior, but its mode lies far outside the range the sampler can represent; for the oscillator, where the exact Kalman likelihood locates it, at $\log\alpha \approx -42$.
PMMH moves toward it until the Euler--Maruyama recursion becomes numerically unstable; from then on, the particle filter returns non-finite likelihoods for a large fraction of the proposals, and only a handful of the $20{,}000$ post-adaptation proposals are accepted.
The second kind are B-PINN runs whose chains have not mixed ($\hat{R} > 1.05$).
For both kinds, we report the median of the draws, which describes where the sampler stopped rather than the posterior.
Marginal plots show such runs as an arrow at the edge of the axis, labelled with the value reached.

\subsubsection{Validation of the Reference Posterior} \label{sec:Exp-Validation}

All biases in this paper are measured against the chain posterior that our method samples.
Three checks confirm that this reference is correct: the first tests the likelihood estimator, the second the simulation of the data, and the third the sampler.

\paragraph{Particle filter against the exact likelihood.}
For the linear oscillator, the Kalman filter gives the exact log-likelihood $\log p(y\mid\lambda)$.
At three points along the drift-invariant direction, the particle-filter estimate deviates from it by $+0.14$, $-0.10$ and $+0.27$\,nats, against Monte Carlo standard deviations of $0.27$, $0.38$ and $0.38$ across five independent filter runs.
The likelihood estimator underlying PMMH thus shows no bias at the resolution relevant for our comparisons.

\paragraph{Simulation step.}
Because the data are simulated with the same scheme that the inference uses, the chain model could benefit from a matched discretization.
To rule this out, we regenerated ten datasets of the oscillator configuration at a $32\times$ finer step and computed the exact (Kalman) chain posterior on each.
Its displacement from the truth is $-0.01 \pm 0.07$ in $\log(c/m)$ and $+0.04 \pm 0.02$ in $\log(k/m)$ (mean $\pm$ standard error over datasets), and its $90\%$ credible intervals contain the truth in $8$ or $9$ of the $10$ datasets for every identified parameter.

\paragraph{A different sampler.}
On the double pendulum and the relativistic particle, we also sampled both models on the state-space latent with NUTS instead of PMMH.
NUTS runs jointly over $(z_0, \xi, \lambda)$, where $\xi = (\xi_1, \dots, \xi_T)$ are the standardized noise increments of Eq.~\eqref{eq:EM}.
The path $z_{0:T}$ is a deterministic function of these variables, and the standard normal prior on $\xi$ does not depend on $\lambda$.
In this non-centred parameterization, the chain model therefore needs no explicit normalizer, while the collider differs from it by $\Delta(\lambda)$ in~\eqref{eq:Metric-Tilt-Exact}.
The resulting posteriors agree with those obtained with PMMH, and the collapse of the collider under torque residuals is reproduced; neither result is therefore an artefact of the particle filter.
NUTS is thus a valid alternative to PMMH for the chain model, but a more expensive one on fine time grids: per effective sample of the drift-invariant scale, it takes $4.9$\,s against $2.7$\,s for PMMH on the pendulum ($1{,}235$ steps), and about $1{,}800$\,s against $6$\,s on the relativistic particle ($12{,}160$ steps; $51$\,h of NUTS in total).

\subsubsection{Full Posterior Results} \label{sec:Exp-FullGrid}

Tables~\ref{tab:grid-osc-sigfix}--\ref{tab:grid-rp-sigfix} report the posterior of every parameter for every inference variant, including the parameters and parameter combinations that Table~\ref{tab:grid} omits.
They show the same pattern as Table~\ref{tab:grid}.
The chain model keeps the drift-invariant scale near the truth and recovers every identified quantity to within $1.3$ posterior standard deviations.
The errors of the collider, in contrast, concentrate where Eq.~\eqref{eq:TwoBiasesDecomposition} predicts: under force residuals on the parameters that enter the residual metric, and on the parameters that enter the dissipation term (the damping of the oscillator, the light speed of the relativistic particle).

\input{Tables/sigma_fixed_tables}

\subsubsection{Replication over Datasets} \label{sec:Exp-Replication}

All runs above use a single dataset per system.
On one dataset, a posterior can lie away from the truth by chance; a bias means that it does so systematically, on average over datasets.
To distinguish the two, we repeat the oscillator configuration of Fig.~\ref{fig:Exp_Osc_Posteriors} (acceleration residuals) on independent datasets, each with a new SDE path and new observation noise.
We do this in two complementary ways.

\paragraph{Sampled posteriors.}
We rerun the B-PINN (NUTS) and our chain model (state space, PMMH) with the settings of Appendix~\ref{sec:Exp-Setup} on ten datasets, including the one of Fig.~\ref{fig:Exp_Osc_Posteriors}.
All B-PINN runs reach $\hat{R} \le 1.001$.

\paragraph{Exact posteriors.}
To remove sampling error and convergence issues altogether, we use ten further datasets and replace the neural network by a surrogate on which all posteriors can be computed exactly.
The collider is a B-PINN on a linear cosine basis ($65$ coefficients with prior standard deviation $0.3$, $Nd = 128$ collocation residuals).
Because the residual is linear in the coefficients, they can be integrated out in closed form, and $Z_\lambda$ can be restored exactly on the same basis, which gives a chain model on the surrogate.
The exact Kalman likelihood of the Euler--Maruyama latent gives the state-space chain.
We evaluate all three posteriors on a grid over the identified coordinates $(\log(c/m), \log(k/m))$, with the log-normal prior of Appendix~\ref{sec:Exp-Setup}; less than $10^{-3}$ of each posterior's mass lies at the boundary of the grid.

\begin{table}[h]
\caption{\textbf{Replication of the oscillator configuration over datasets} (acceleration residuals). Top: the neural B-PINN and our chain model, sampled as in the main text, on ten datasets. Bottom: exactly computed posteriors on a linear surrogate, on ten further datasets. \emph{displ.}: mean over datasets of the posterior mean of the log parameter minus the log truth, $\pm$ its standard error. \emph{sd}: mean posterior standard deviation. \emph{$90\%$ cov.}: number of datasets whose central $90\%$ credible interval contains the truth.}
\label{tab:replication}
\vskip 0.05in
\begin{center}
\begin{small}
\setlength{\tabcolsep}{3pt}
\begin{tabular}{@{}l rrr rrr@{}}
\toprule
 & \multicolumn{3}{c}{$\log(c/m)$} & \multicolumn{3}{c}{$\log(k/m)$} \\
\cmidrule(lr){2-4} \cmidrule(lr){5-7}
\textbf{Model (latent)} & displ. & sd & $90\%$ cov. & displ. & sd & $90\%$ cov. \\
\midrule
\multicolumn{7}{@{}l}{\textit{Sampled posteriors, ten datasets}} \\
\addlinespace[2pt]
Chain (state space) & $+0.02 \pm 0.05$ & $0.17$ & $10/10$ & $+0.02 \pm 0.02$ & $0.11$ & $10/10$ \\
Collider B-PINN (MLP) & $\mathbf{-0.42 \pm 0.03}$ & $0.24$ & $\mathbf{4/10}$ & $\mathbf{-0.12 \pm 0.03}$ & $0.11$ & $\mathbf{7/10}$ \\
\midrule
\multicolumn{7}{@{}l}{\textit{Exact posteriors on a linear surrogate, ten datasets}} \\
\addlinespace[2pt]
Chain (state space) & $-0.03 \pm 0.05$ & $0.17$ & $8/10$ & $+0.02 \pm 0.04$ & $0.11$ & $9/10$ \\
Chain (linear surrogate) & $-0.05 \pm 0.05$ & $0.18$ & $9/10$ & $+0.02 \pm 0.03$ & $0.09$ & $9/10$ \\
Collider (linear surrogate) & $\mathbf{-0.46 \pm 0.04}$ & $0.23$ & $\mathbf{1/10}$ & $\mathbf{-0.08 \pm 0.03}$ & $0.10$ & $\mathbf{7/10}$ \\
\bottomrule
\end{tabular}
\end{small}
\end{center}
\end{table}

\paragraph{Results.}
Table~\ref{tab:replication} confirms that the displacements reported in the main text are a systematic bias.
Over ten datasets, the B-PINN underestimates $c/m$ by $0.42 \pm 0.03$ log units and $k/m$ by $0.12 \pm 0.03$, and it lies below the truth on every dataset in both coordinates.
Measured relative to the chain model on the same dataset, which removes the variation between datasets, the displacements are $-0.45 \pm 0.02$ and $-0.14 \pm 0.02$.
The B-PINN posterior is not narrower than the chain's (mean standard deviation $0.24$ against $0.17$ in $\log(c/m)$), but it is displaced by almost two of its own standard deviations, so its $90\%$ credible intervals contain the truth in only $4$ and $7$ of the ten datasets, whereas those of the chain model contain it in all ten.
The dataset of Fig.~\ref{fig:Exp_Osc_Posteriors} lies at the upper end: its displacement in $\log(c/m)$ ($-0.53$) is the third largest of the ten, against a median of $-0.39$.

The exact posteriors on the linear surrogate show the same picture, with displacements of $-0.46 \pm 0.04$ and $-0.08 \pm 0.03$ and coverage in $1$ and $7$ of ten datasets.
Restoring $Z_\lambda$ exactly on the same surrogate removes the bias in both coordinates and restores nominal coverage, as does the state-space chain.
The bias is therefore reproducible across datasets, and it is caused by the missing normalizer rather than by the particular dataset or by the limited expressiveness of the surrogate.


\subsubsection{Sensitivity to the Residual Scale} \label{sec:Exp-SigmaSweep}

All experiments above fix $\sigma_r$ at the value that generated the data.
To test how the bias depends on this value, we repeat the exact-posterior comparison of Appendix~\ref{sec:Exp-Replication} on the same ten oscillator datasets (acceleration residuals), varying $\sigma_r$ in the data and in the models.
For $\sigma_r \le 0.1$, we evaluate the posteriors on a finer grid, since they are narrower than its original spacing.

\begin{table}[h]
\caption{\textbf{Sensitivity to the residual scale} (oscillator, acceleration residuals, exact posteriors, ten datasets). Displacement of $\log(c/m)$ from the truth (mean over datasets $\pm$ standard error) and number of datasets whose central $90\%$ credible interval contains the truth. Top: $\sigma_r$ matched to the data. Bottom: $\sigma_r$ misspecified.}
\label{tab:sigma-sweep}
\vskip 0.05in
\begin{center}
\begin{small}
\setlength{\tabcolsep}{4pt}
\begin{tabular}{@{}cc rr rr rr@{}}
\toprule
\multicolumn{2}{c}{$\sigma_r$} & \multicolumn{2}{c}{Chain (state space)} & \multicolumn{2}{c}{Chain (linear surrogate)} & \multicolumn{2}{c}{Collider (linear surrogate)} \\
\cmidrule(lr){1-2} \cmidrule(lr){3-4} \cmidrule(lr){5-6} \cmidrule(lr){7-8}
data & model & displ. & $90\%$ cov. & displ. & $90\%$ cov. & displ. & $90\%$ cov. \\
\midrule
$0.05$ & $0.05$ & $-0.01 \pm 0.02$ & $9/10$ & $-0.03 \pm 0.02$ & $9/10$ & $\mathbf{-0.07 \pm 0.02}$ & $\mathbf{7/10}$ \\
$0.1$  & $0.1$  & $-0.01 \pm 0.03$ & $9/10$ & $-0.02 \pm 0.03$ & $9/10$ & $\mathbf{-0.12 \pm 0.03}$ & $\mathbf{7/10}$ \\
$0.3$  & $0.3$  & $-0.03 \pm 0.05$ & $8/10$ & $-0.05 \pm 0.05$ & $9/10$ & $\mathbf{-0.46 \pm 0.04}$ & $\mathbf{1/10}$ \\
$1.0$  & $1.0$  & $-0.04 \pm 0.07$ & $8/10$ & $-0.16 \pm 0.07$ & $9/10$ & $\mathbf{-0.74 \pm 0.03}$ & $\mathbf{0/10}$ \\
\midrule
$0$    & $0.3$  & $\mathbf{+0.40 \pm 0.01}$ & $\mathbf{0/10}$ & $\mathbf{+0.37 \pm 0.01}$ & $\mathbf{1/10}$ & $-0.15 \pm 0.01$ & $10/10$ \\
$0.3$  & $0.9$  & $\mathbf{+0.97 \pm 0.05}$ & $\mathbf{0/10}$ & $\mathbf{+0.96 \pm 0.05}$ & $\mathbf{0/10}$ & $-0.30 \pm 0.02$ & $10/10$ \\
\bottomrule
\end{tabular}
\end{small}
\end{center}
\end{table}

With $\sigma_r$ matched to the data, the collider bias is present at every scale and grows with it, from $-0.07$ to $-0.74$ in $\log(c/m)$, while both chain models remain close to the truth (Table~\ref{tab:sigma-sweep}, top).
At small $\sigma_r$, the displacement is small in absolute terms but still about one posterior standard deviation, and coverage remains below nominal.

A misspecified residual scale introduces a different bias, which affects the chain model as well.
For noise-free data, or for a residual scale three times too large, the chain overestimates the damping, and its intervals miss the truth (Table~\ref{tab:sigma-sweep}, bottom).
The downward collider bias then partly offsets this error, so that the collider's wider intervals cover the truth; this reflects two biases of opposite sign cancelling, not a correct model.

Ideally, the residual scale is therefore matched to the data.
When it is unknown, a natural remedy is to infer $\sigma_r$ jointly with $\lambda$, as noted in Section~\ref{sec:Conclusions}.
For the chain model, this is a direct extension of PMMH, since the per-step normalizers~\eqref{eq:Z-StateSpace} remain analytic in $\sigma_r$.
For the collider, however, the bias $\Delta$ then also depends on $\sigma_r$, which gives rise to a further bias mechanism; we leave its analysis to future work.

\subsubsection{The Dissipation Term on the Conservative Systems} \label{sec:Exp-Conservative}

The double pendulum and the relativistic particle are both conservative, yet the dissipation term of a global surrogate need not vanish for them.
We first explain why, and then evaluate the term for each system.

\paragraph{Why conservative systems can retain a dissipation term.}
Suppose that the forces derive from a (possibly time-dependent) potential and that $M_\lambda = \partial^2\mathcal{L}/\partial v\,\partial v^\top$ is the velocity Hessian of the Lagrangian $\mathcal{L}$.
Then the flow is Hamiltonian in the canonical coordinates $(x, p)$ with $p = \partial\mathcal{L}/\partial v$, and divergence-free in these coordinates (Liouville's theorem).
Since the change of variables $(x, v) \mapsto (x, p)$ has Jacobian determinant $\det M_\lambda$,
\begin{equation} \label{eq:Model_Difference_OnsagerMachlup}
    \nabla\!\cdot\! f_\lambda(z(t),t) = -\frac{\mathrm{d}}{\mathrm{d}t}\log\det M_\lambda(z(t)), \qquad
    \int_0^{\mathcal{T}} \nabla\!\cdot\! f_\lambda\,\mathrm{d}t = \log\frac{\det M_\lambda(z(0))}{\det M_\lambda(z(\mathcal{T}))},
\end{equation}
a boundary term that does not grow with $\mathcal{T}$.
On a global surrogate, however, $\Delta(\lambda)$ depends on $\sum_k|\mathrm{Re}\,s_k|$ (Eq.~\eqref{eq:Tilt-Diss-General}) rather than on the divergence, and the two can differ in two ways.
First, the eigenvalues of a Hamiltonian linearization come in pairs $\pm s$: oscillatory pairs ($\mathrm{Re}\,s = 0$) contribute nothing, but hyperbolic pairs (real $\pm s$) contribute $\mathcal{T}|s|$ although their divergence vanishes, so a chaotic system can retain a $\lambda$-dependent term that grows with $\mathcal{T}$.
Second, if $Df_\lambda$ has a single eigenvalue with nonzero real part, as for a one-dimensional system with velocity-dependent inertia, then $\sum_k|\mathrm{Re}\,s_k| = |\nabla\!\cdot\! f_\lambda|$ pointwise, and the windowed dissipation term becomes $-\frac{1}{2}\int_0^{\mathcal{T}} |\nabla\!\cdot\! f_\lambda|\,\mathrm{d}t$.
This grows linearly in $\mathcal{T}$ whenever the divergence changes sign along the orbit, although $\int_0^{\mathcal{T}} \nabla\!\cdot\! f_\lambda\,\mathrm{d}t$ stays bounded.
The pendulum is an instance of the first case and the relativistic particle of the second.
We evaluate both quantities along the noise-free trajectory at the ground truth; this explains why the dissipation term leaves the pendulum results in Table~\ref{tab:grid} unaffected but biases the relativistic particle.

\paragraph{Double pendulum.}
The time average of $\sum_k|\mathrm{Re}\,s_k|$ over $[0, \mathcal{T}]$ is $0.508$, so the surrogate retains a term of $\tfrac{\mathcal{T}}{2}\langle\sum_k|\mathrm{Re}\,s_k|\rangle = 3.05$\,nats at $\mathcal{T} = 12$, although the mean divergence is only $1.2\cdot10^{-2}$.
This term does not affect the shift $\log\alpha$ along the drift-invariant direction reported in Table~\ref{tab:grid}: scaling $(m_1, m_2) \to \alpha(m_1, m_2)$ leaves the acceleration $a_\lambda = M_\lambda^{-1}F_\lambda$, and hence $Df_\lambda$, unchanged, and indeed the time average agrees to six digits at $\alpha = 0.5$, $1$ and $2$.
Off the direction, it is negligible as well: it varies by less than $2\cdot10^{-3}$ over $\pm 0.05$ in $\log \ell_1$, i.e., $|\partial\Delta/\partial\log\ell_1| \approx 0.04$\,nats per unit of $\log\ell_1$, which is small against a prior with log-standard deviation $0.25$.

\paragraph{Relativistic particle.}
The drive returns the particle to rest at both ends of the record ($\gamma(0) = \gamma(\mathcal{T}) = 1$), so $\int_0^{\mathcal{T}}\nabla\!\cdot\! f_\lambda\,\mathrm{d}t = 0$ to machine precision: the divergence term is not merely bounded but absent.
The drift Jacobian $Df_\lambda = \begin{bsmallmatrix} 0 & 1 \\ 0 & \nabla\cdot f_\lambda\end{bsmallmatrix}$, however, has the eigenvalues $0$ and $\nabla\!\cdot\! f_\lambda = -3 q E(t) v / (m_0 c^2 \gamma)$, which changes sign twice per drive period.
Hence $\sum_k|\mathrm{Re}\,s_k| = |\nabla\!\cdot\! f_\lambda|$, and $\int_0^{\mathcal{T}}|\nabla\!\cdot\! f_\lambda|\,\mathrm{d}t = 10.36$, so that $\Delta_{\mathrm{diss}} = -5.18$\,nats at the truth, a term that grows linearly in $\mathcal{T}$.
Differentiating with respect to the light speed at a fixed trajectory gives $\partial\Delta_{\mathrm{diss}}/\partial\log c = +9.5$\,nats per unit of $\log c$.
The collider is thus rewarded for a larger $c$, because a larger $c$ keeps $\gamma$ closer to one and thereby flattens the divergence.
This reward has the sign and the order of magnitude of the observed error, and it is too strong for a prior with log-standard deviation $0.25$ to counteract.
The B-PINN reports $c = 1.78$ ($\log c$ displaced by $+0.569$), against $0.89$ for the chain.
We do not derive a quantitative prediction from these numbers, because $\Delta_{\mathrm{diss}}$ is strongly nonlinear in $c$ (roughly $\propto -c^{-2}$) over the range that the posterior covers.

\subsubsection{Restoring the Normalizer on the Neural Surrogate} \label{sec:Exp-NetChain}

In principle, the bias could stem from the probabilistic model (collider or chain), the latent (state space or neural surrogate), or the estimator (PMMH or NUTS).
On the state-space latent, the model alone decides: chain and collider share latent and estimator and coincide exactly whenever $\Delta(\lambda)$ is constant (Section~\ref{sec:Exp-Grid}), and NUTS reproduces the PMMH results (Appendix~\ref{sec:Exp-Validation}).
Here, we hold the neural surrogate fixed and change only the model.

\paragraph{Construction.}
Chain (Fourier MLP) adds the term $-\Delta = \log Z_r - \log Z_\lambda$ to the log-density of the oscillator B-PINN.
By Eq.~\eqref{eq:Model_Difference}, this turns the collider into the chain, up to the approximations of Appendix~\ref{sec:Tilt-Derivation}.
For the oscillator, Eq.~\eqref{eq:TwoBiasesDecomposition} at $d = 1$, with $\det A_\lambda = m$ under force residuals and $\det A_\lambda = 1$ under acceleration residuals, gives
\begin{equation} \label{eq:NetChain-Factor}
    \log Z_r - \log Z_\lambda = \underbrace{\frac{c\,\mathcal{T}}{2m}}_{\text{dissipation}} \; + \; \underbrace{n_{\mathrm{eff}}\log m}_{\text{metric, force residuals only}} + \mathrm{const}.
\end{equation}
Everything else (architecture, weight prior, data, collocation grid, sampler and seed) is identical to the B-PINN.

\paragraph{Effective number of constraints.}
The correction requires $n_{\mathrm{eff}} = \operatorname{tr}[G_\lambda(G_\lambda + P)^{-1}]$.
We compute it once, from the Gauss--Newton curvature at $\lambda = \lambda_{\mathrm{true}}$, and keep it fixed during sampling.
This is the idealization under which the closed forms are derived (assumption~(iii) in Appendix~\ref{sec:Tilt-Derivation}): by Eq.~\eqref{eq:Tilt-Neff-Spectrum}, $n_{\mathrm{eff}}$ depends on $\lambda$ only through $\kappa_\lambda$ and is constant in the saturated regime.
We evaluate the curvature at the final network weights of the corresponding B-PINN run rather than at the data-free mode $\theta^* = 0$, where $x_\theta \equiv 0$ solves the ODE but the network Jacobian is degenerate (assumption~(ii)).
With $Nd = 128$ and $\dim\theta = 1441$, this gives $n_{\mathrm{eff}} = 110.2$ for force residuals.
For acceleration residuals, only the dissipation term is active and no $n_{\mathrm{eff}}$ is needed.

This value depends only moderately on $\lambda$ but strongly on the network weights.
Evaluating the curvature at the chain's posterior mean instead of $\lambda_{\mathrm{true}}$ changes it by less than $0.5\%$, and over draws from the chain posterior it varies by about $\pm 10\%$ ($98$--$116$ around $110.2$).
At the final state of the acceleration-residual B-PINN, by contrast, it is $99.9$, and at a MAP fit of the network it drops to $44$.
Moreover, $n_{\mathrm{eff}}$ is not constant along the B-PINN's own posterior ($55$--$78$, where the B-PINN has collapsed the mass scale), so the saturated regime of assumption~(iii) holds only approximately.
This sensitivity to the expansion point is the main source of the overcorrection discussed below.

\begin{table}[h]
\caption{\textbf{Restoring the normalizer on the neural surrogate (oscillator).} Displacements from the truth in log space (posterior mean of the log parameter minus the log truth) for the Collider B-PINN and for Chain (Fourier MLP), and the fraction of the displacement removed by the correction, $1 - |\text{chain}|/|\text{collider}|$. \emph{Active terms} are the terms of Eq.~\eqref{eq:NetChain-Factor} that depend on the inferred parameters in that configuration. The last configuration is the $\mathcal{T} = 24$ run of the horizon study in Fig.~\ref{fig:Exp_Osc_Horizon}.}
\label{tab:netchain}
\vskip 0.05in
\begin{center}
\begin{small}
\begin{tabular}{@{}lllrrr@{}}
\toprule
\textbf{Configuration} & \textbf{Active terms} & \textbf{Direction} & \textbf{Collider} & \textbf{Chain (MLP)} & \textbf{Removed} \\
\midrule
force residuals & metric, dissipation & $\log m$ & $-1.453$ & $+0.497$ & $65.8\%$ \\
\addlinespace
acceleration residuals & dissipation & $\log (c/m)$ & $-0.526$ & $-0.013$ & $97.5\%$ \\
\bottomrule
\end{tabular}
\end{small}
\end{center}
\end{table}

\paragraph{Results.}
Table~\ref{tab:netchain} reports two configurations, each of which activates a different subset of the terms in~\eqref{eq:NetChain-Factor}.
All runs converge ($\hat{R} = 1.000$ over $24{,}000$ draws), so these are valid posteriors.
How much of the bias the correction removes follows how accurate the closed forms are in each configuration:
\begin{itemize}
    \item With acceleration residuals, only the dissipation term is active and no $n_{\mathrm{eff}}$ enters. The correction removes $97.5\%$ of the displacement of $\log(c/m)$.
    \item With force residuals, the approximate $n_{\mathrm{eff}}$ matters, and the correction can overshoot. The collider's $\log m = -1.45$ becomes $+0.50$ ($m \approx 1.64$ against a truth of $1.00$): the correction crosses the truth and leaves a displacement of about one third of the original on the other side. A term linear in $\log m$ with a fixed coefficient $n_{\mathrm{eff}}$ replaces a function of $\kappa_\lambda$ and can therefore overcorrect.
\end{itemize}
As cautioned at the end of Appendix~\ref{sec:Tilt-Derivation}, the closed forms thus locate the bias and remove most of it, but they do not calibrate it exactly.

\paragraph{Remaining stiffness bias.}
The correction also leaves part of the stiffness bias: with acceleration residuals, Chain (Fourier MLP) reports $k/m = 3.48$, against $3.16$ for the B-PINN and $3.93$ for the chain on the state-space latent (truth $4.00$).
This remainder is still collider bias rather than a representation error of the surrogate.
On the linear cosine surrogate of Appendix~\ref{sec:Exp-Replication}, where $Z_\lambda$ can be restored exactly, the collider displaces $\log(k/m)$ from the truth by $-0.08 \pm 0.03$ (mean $\pm$ standard error over ten datasets), whereas the exactly restored chain ($+0.02 \pm 0.03$) and the state-space chain ($+0.02 \pm 0.04$) are unbiased.
The leading-order correction misses this part because $k$ enters $\log Z_\lambda$ only through the $\mathcal{O}(\beta/\Omega)$ remainder of Eq.~\eqref{eq:Tilt-dIdbeta}.

\subsubsection{Details on the Scaling-Law Experiments} \label{sec:Exp-LawDetail}

Figs.~\ref{fig:Exp_Osc_Horizon} and~\ref{fig:Exp_Osc_Exact} use the oscillator configuration of Appendix~\ref{sec:Exp-Setup}, in each case with the mechanism that is not under test switched off.
Fig.~\ref{fig:Exp_Osc_Exact} checks the metric term of $\Delta(\lambda)$ in a setting where its closed form is exact; Fig.~\ref{fig:Exp_Osc_Horizon} tests the approximate dissipation term on the neural surrogate.

\begin{figure}[t]
    \centering
    \includegraphics[width=0.5\textwidth]{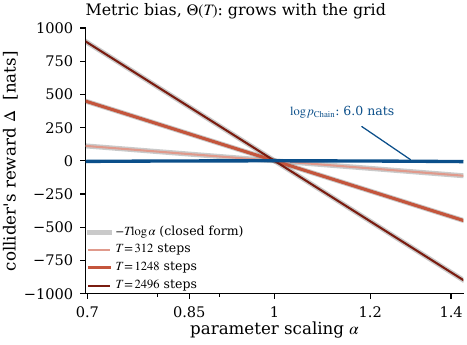}
    \caption{\textbf{Exact closed form of the residual metric bias against measurements on the oscillator} (Eq.~\eqref{eq:Metric-Tilt-Exact}). Closed form in grey, the collider on the Markovian state-space latent in dark red, the chain in blue. On the state-space latent, $\Delta(\lambda)$ along the drift-invariant direction matches $-T\log\alpha$ to within $2.8 \cdot 10^{-12}$\,nats for each number of time steps shown ($T = 312$, $1{,}248$, $2{,}496$; $T = 624$, omitted for clarity, agrees equally), whereas the chain's profile stays flat.}
    \label{fig:Exp_Osc_Exact}
\end{figure}

\paragraph{Residual metric bias (Fig.~\ref{fig:Exp_Osc_Exact}).}
The figure shows $\Delta(\lambda)$ along the drift-invariant direction $(m,c,k) \to \alpha(m,c,k)$ on the state-space latent, for force residuals; the Markovian latent switches off the dissipation term.
We use four discretizations, $T \in \{312, 624, 1248, 2496\}$.
The measured $\Delta(\lambda)$ matches $-T\log\alpha$ to within $2.8\cdot10^{-12}$\,nats, i.e., to machine precision, as expected since Eq.~\eqref{eq:Metric-Tilt-Exact} is exact on this latent.
Over the swept range of $\alpha$, $\Delta(\lambda)$ varies by $225$, $449$, $899$ and $1797$\,nats, doubling with the number of steps, whereas the chain's log-posterior varies by only $6.0$\,nats over the same range.

\paragraph{Dissipation bias (Fig.~\ref{fig:Exp_Osc_Horizon}).}
Unlike the metric term, the dissipation term is not a property of the latent alone, so we measure its effect on the posterior rather than $\Delta(\lambda)$ itself.
We use acceleration residuals, which switch off the metric bias, and vary the horizon over $\mathcal{T} \in \{12, 24, 48\}$; the run at $\mathcal{T} = 24$ is the standard oscillator configuration.
Across horizons, we keep the surrogate's bandwidth per unit time and the Euler step fixed and scale $K$, $N$ and $T$ with $\mathcal{T}$, so that only the amount of data changes.
The panel shows the posterior mean of $\log(c/m)$ minus its true value for Chain (state space), Chain (Fourier MLP) and the Collider B-PINN, each with a bar of $\pm 1$ posterior standard deviation.
The state-space collider is not plotted separately, since under acceleration residuals $\Delta(\lambda)$ is constant and its draws are identical to the chain's.

The B-PINN lies at $-0.377$, $-0.526$ and $-0.669$, i.e., at $1.55$, $2.08$ and $2.67$ of its own posterior standard deviations, which barely change ($0.243$, $0.253$, $0.251$).
The chain stays near the truth, both on the state-space latent ($-0.086$, $-0.022$, $+0.041$, with standard deviations $0.198$, $0.176$, $0.143$) and on the surrogate with the correction ($-0.095$, $-0.013$, $+0.141$).
The chain's posterior thus contracts as the record grows, whereas the B-PINN's does not, so the B-PINN's error grows relative to its own stated uncertainty.

To compare with the closed form~\eqref{eq:Tilt-Osc-Delta}, we reweight the chain posterior of $\log(c/m)$, approximated as Gaussian, by $e^{\Delta_{\mathrm{diss}}} = e^{-(c/m)\mathcal{T}/2}$.
This predicts shifts relative to the chain of $-0.21$, $-0.32$ and $-0.41$, whereas the B-PINN is shifted by $-0.29$, $-0.50$ and $-0.71$.
The leading-order term thus predicts the direction of the bias and its growth with the record, but only between three quarters and three fifths of its size; it also predicts a posterior that contracts with $\mathcal{T}$, which the B-PINN's does not.

 
\subsection{Comparison with Nested Inference} \label{sec:BilionisComparison}

To our knowledge, the only prior work that adopts the chain model, by Alberts, Hao, and Bilionis~\citep{Bilionis1, Bilionis2, Bilionis3}, handles the unknown normalizer $Z_\lambda$ by nested inference: an inner loop estimates the gradient $\partial_\lambda \log Z_\lambda$ anew at every step of an outer loop over the joint state.
We compare our method, the particle filter with PMMH (Section~\ref{sec:InferenceScheme}), against the two nested schemes that target the same chain posterior.
\begin{itemize}
    \item \textbf{NSVI}~\citep{Bilionis2} (nested stochastic variational inference) fits an outer Gaussian guide over path coefficients, physical parameters and initial state by maximizing an ELBO. The normalizer gradient in this ELBO is supplied by a second, persistent inner guide, which is re-fitted at every outer step.
    \item \textbf{NPSGLD}~\citep{Bilionis3} (nested preconditioned stochastic-gradient Langevin dynamics) replaces both guides by preconditioned Langevin chains: a persistent auxiliary chain over the path supplies the normalizer gradient to a main chain over the joint state.
\end{itemize}
We use the public JAX reimplementation of both methods by \citet{Bolt2026_NIFFReplication} (commit \texttt{dab4403}), which is not by the original authors; we run it unmodified and supply only the problem definition and the evaluation.
To isolate the inference algorithm, all methods use the Euler--Maruyama path as latent, on which $Z_\lambda$ is in fact available in closed form.
The comparison therefore measures how expensive and how accurate nested estimation of $Z_\lambda$ is where it could be avoided.
The baselines were designed for latents without a closed-form normalizer, a setting we do not test.

\paragraph{Setup.}
We use two systems in the configurations of Section~\ref{sec:Exp-Systems}: the damped oscillator under both residual metrics, and the double pendulum under torque residuals.
All methods share the model, the latent ($T = 1248$ Euler--Maruyama steps for the oscillator and $T = 1235$ for the pendulum), the observations, and the log-normal priors with median at the truth.
As in all experiments, $\sigma_r$ is fixed at its true value.
Instead of tuning a single run per baseline, we sweep their hyperparameters.
For NSVI, we vary the outer learning rate in $[5\cdot10^{-5}, 10^{-3}]$ and the number of inner iterations from $8$ to $256$, with budgets of up to $1.2\cdot10^{6}$ iterations.
For NPSGLD, we vary the step size in $[10^{-8}, 10^{-3}]$, use $4$ or $16$ chains and both of its preconditioners (\texttt{rmsprop} and \texttt{dense\_fisher}), and switch step-size annealing on or off, with budgets of up to $10^{7}$ iterations.
This amounts to $23$ runs on the oscillator and $21$ on the pendulum.
All runs, including those of our method, use the same Quadro RTX 8000 GPU, so that run times are directly comparable.

\paragraph{Evaluation metrics.}
Wall-clock time alone favours samplers that have not mixed, since they produce many draws that carry little information.
We therefore measure cost in \emph{seconds per effective sample}, i.e., wall-clock time divided by the effective sample size (ESS), and report the split-$\hat{R}$ alongside it.
As in Table~\ref{tab:grid}, runs with $\hat{R} > 1.05$ do not yield a valid posterior; we report their median and mark them with a dagger.
For our method, the ESS is the smallest over the components of $\lambda$; for NPSGLD, it is the ESS of the drift-invariant scale.
This choice favours the baseline: on the drift-invariant scale alone, the ESS of our method on the pendulum is about $870$ rather than $343$, which would lower its cost from $6.71$ to $2.6$\,s per effective sample.
The cost and $\hat{R}$ of our method refer to a single PMMH chain.
To check its mixing, we additionally ran four independent chains (sampler seeds $0$--$3$) per configuration; they give split-$\hat{R} \le 1.001$ for all parameters, and their posterior means of the drift-invariant scale agree to within $0.02$ (oscillator) and $0.003$ (pendulum).
NSVI draws independent samples from a fitted guide, so $\hat{R}$ and ESS do not apply to it; we assess it by its bias instead.
Finally, the \emph{width} of a run is the ratio of its posterior standard deviation to that of our method.
It measures what matters for uncertainty quantification: a method can locate the posterior correctly and still misrepresent its uncertainty.

\begin{table}[h]
\caption{\textbf{Nested inference compared with our particle filter with PMMH.} Ranges are over the hyperparameter sweep of each method. \emph{Location} is the drift-invariant scale ($m$ for the oscillator, $m_1{+}m_2$ for the pendulum), reported as posterior mean $\pm$ standard deviation where $\hat{R} \le 1.05$ and as a daggered median otherwise. \emph{Width} is the posterior standard deviation divided by that of our method; $1.00$ is correct, and $0.03$ means thirty-fold overconfidence. \emph{Cost} is in seconds per effective sample. A run counts as \emph{diverged} if its location is off by more than a factor of $100$ or its spread exceeds $100$ reference standard deviations.}
\label{tab:bilionis}
\vskip 0.05in
\begin{center}
\begin{small}
\begin{tabular}{@{}llrlrrlr@{}}
\toprule
\textbf{System} & \textbf{Method} & \textbf{Runs} & \textbf{Location} & $\hat{R}$ & \textbf{ESS} & \textbf{Width} & \textbf{Cost} \\
\midrule
Oscillator & PF\,+\,PMMH (ours) & $1$ & $1.015 \pm 0.140$ & $1.00$ & $604$ & $1.00$ & $\mathbf{0.81}$ \\
force residuals & NSVI & $8$ & $0.78$--$1.43$ & --- & --- & $0.15$--$0.28$ & --- \\
 & NPSGLD & $10$ & $1.00$--$1.12$\,$\dagger$ & $1.3$--$3.0$ & $3$--$32$ & $0.03$--$1.25$ & $56$--$2385$ \\
\addlinespace
Oscillator & PF\,+\,PMMH (ours) & $1$ & $1.035 \pm 0.220$ & $1.00$ & $875$ & $1.00$ & $\mathbf{0.51}$ \\
accel.\ residuals & NSVI & $2$ & $1.28$ & --- & --- & $1.36$--$1.38$ & --- \\
 & NPSGLD & $3$ & $1.00$--$1.12$\,$\dagger$ & $1.1$--$1.5$ & $6$--$86$ & $1.09$--$1.90$ & $21$--$190$ \\
\addlinespace
Double pendulum & PF\,+\,PMMH (ours) & $1$ & $1.615 \pm 0.394$ & $1.00$ & $343$ & $1.00$ & $\mathbf{6.71}$ \\
torque residuals & NSVI & $16$ & $0.17$--$1.58$ & --- & --- & $0.03$--$0.10$ & --- \\
 & NPSGLD & $5$ & $1.50$--$1.92$\,$\dagger$, $2$ diverged & $2.5$--$277$ & $3$--$4$ & $0.07$--$0.94$ & $141$--$495$ \\
\bottomrule
\end{tabular}
\end{small}
\end{center}
\end{table}

\paragraph{Cost.}
Our method produces one effective sample every $0.5$--$6.7$\,s.
NPSGLD, the only baseline that produces MCMC samples, needs $21$ to $2{,}385$\,s per effective sample, i.e., $21\times$ to $2{,}900\times$ more, depending on the configuration.
Its cheapest pendulum run finishes about five times sooner than our method, but it is $21\times$ more expensive per effective sample, since its $423$\,s yield only three effective samples.
A larger budget does not improve its mixing either: the longest NPSGLD run, with $10^{7}$ iterations over $11{,}444$\,s, yields five effective samples.

\paragraph{NSVI converges to a biased posterior.}
On the oscillator, NSVI is fast, and most of its settings agree on the drift-invariant scale; but the posterior they agree on is the wrong one.
In six of its eight force-residual runs, the damping ratio lies at $c/m = 0.78$--$0.81$, whereas the truth is $1.20$ and our method gives $1.209 \pm 0.236$, an underestimate of about $1.8$ reference standard deviations.
The other two runs, both with $10^{5}$ outer iterations, end at $c/m = 0.69$ and $0.94$.
More computation does not reduce the bias: raising the outer iterations from $3\cdot10^{4}$ to $3\cdot10^{5}$ and the inner iterations from $32$ to $256$ increases the run time $113$-fold, from $128$\,s to $14{,}474$\,s, but moves $c/m$ only from $0.790$ to $0.800$.
The bias is therefore a property of the converged approximation, not of insufficient computation.
On the chaotic pendulum, NSVI does not converge at all: across $16$ settings, the mass scale ranges from $0.17$ to $1.58$, so its final estimate reflects where the budget ran out rather than the posterior.
It is also strongly overconfident there: its posterior standard deviations are only $0.03$--$0.10$ times that of our method (Table~\ref{tab:bilionis}), i.e., ten to thirty times too small, so that its scattered estimates come with misleadingly narrow uncertainty.

\paragraph{NPSGLD misrepresents the uncertainty.}
None of the $18$ NPSGLD runs, on either system and in any configuration, reaches $\hat{R} \le 1.05$.
Because its chains do not mix, its narrow posteriors reflect insufficient exploration rather than precision.
On the pendulum, NPSGLD reports $m_1{+}m_2 = 1.495$ with a standard deviation fourteen times too small; on the oscillator with force residuals, it reports $m = 1.000$ with a standard deviation thirty times too small.
At the same time, its posteriors over the identified ratios are up to $2.5\times$ too wide.
The NPSGLD posterior is therefore not only mislocated but also misshapen.
Our method, in contrast, converges in every configuration ($\hat{R} \le 1.001$ over four independent chains) and is asymptotically exact.

\paragraph{Why the two systems differ.}
Both baselines perform considerably better on the oscillator than on the pendulum, for structural reasons.
The oscillator is linear, so its path posterior is close to Gaussian, and the Gaussian outer guide of NSVI is nearly well specified.
Moreover, with force residuals, the comparatively large process noise ($\sigma_r = 0.30$) determines the drift-invariant scale, which is the direction that nested schemes find most difficult.
The pendulum, in contrast, is chaotic, its mass matrix depends on the state, and its path posterior over $1{,}235$ steps and two degrees of freedom is strongly non-Gaussian.
Nested inference therefore does not fail everywhere, but it succeeds only where the problem is close to what its approximations assume, and even there NPSGLD costs more than $40\times$ as much per effective sample as our method.

\end{document}

%% file: Tables/sigma_fixed_tables.tex
\begin{table}[t]
    \caption{\textbf{Oscillator.} The three inference legs under both residual metrics, on one dataset and one prior, with $\sigma_r$ held at its true value. Posterior mean $\pm$ standard deviation over 14000 draws (state space) and 24000 draws (B-PINN). $\dagger$: a walk at the Euler map's stability boundary, not a posterior; its median is quoted; $\ddagger$: chains that have not merged ($\hat{R} > 1.05$), likewise a median rather than a posterior.}
    \label{tab:grid-osc-sigfix}
    \vskip 0.05in
    \begin{center}
    \begin{small}
    \resizebox{\columnwidth}{!}{%
    \begin{tabular}{@{}llrrrrr@{}}
    \toprule
    \textbf{residual} & \textbf{leg} & $m$ (1.00) & $c$ (1.20) & $k$ (4.00) & $c/m$ (1.20) & $k/m$ (4.00) \\
    \midrule
    force residual & Chain (state space) & $1.015 \pm 0.14$ & $1.209 \pm 0.20$ & $3.981 \pm 0.51$ & $1.209 \pm 0.24$ & $3.948 \pm 0.43$ \\
     & Collider (state space) & $9.3{\cdot}10^{-6}$\,$\ddagger$ & $3.7{\cdot}10^{-4}$\,$\ddagger$ & $0.03387$\,$\ddagger$ & $39.74$\,$\ddagger$ & $3656$\,$\ddagger$ \\
     & Collider (B-PINN) & $0.2391 \pm 0.050$ & $0.5663 \pm 0.13$ & $1.832 \pm 0.25$ & $2.462 \pm 0.73$ & $7.935 \pm 1.7$ \\
    \addlinespace
    acceleration residual & Chain (state space) & $1.035 \pm 0.22$ & $1.219 \pm 0.28$ & $4.030 \pm 0.79$ & $1.191 \pm 0.20$ & $3.927 \pm 0.41$ \\
     & Collider (state space) & $1.035 \pm 0.22$ & $1.219 \pm 0.28$ & $4.030 \pm 0.79$ & $1.191 \pm 0.20$ & $3.927 \pm 0.41$ \\
     & Collider (B-PINN) & $1.314 \pm 0.28$ & $0.9419 \pm 0.24$ & $4.115 \pm 0.81$ & $0.7311 \pm 0.18$ & $3.162 \pm 0.35$ \\
    \bottomrule
    \end{tabular}%
    }
    \end{small}
    \end{center}
\end{table}

\begin{table}[t]
    \caption{\textbf{Double pendulum.} The three inference legs under both residual metrics, on one dataset and one prior, with $\sigma_r$ held at its true value. Posterior mean $\pm$ standard deviation over 14000 draws (state space) and 24000 draws (B-PINN). $\dagger$: a walk at the Euler map's stability boundary, not a posterior; its median is quoted; $\ddagger$: chains that have not merged ($\hat{R} > 1.05$), likewise a median rather than a posterior.}
    \label{tab:grid-dp-sigfix}
    \vskip 0.05in
    \begin{center}
    \begin{small}
    \resizebox{\columnwidth}{!}{%
    \begin{tabular}{@{}llrrrrrr@{}}
    \toprule
    \textbf{residual} & \textbf{leg} & $m_1$ (1.00) & $m_2$ (0.500) & $\ell_1$ (1.00) & $\ell_2$ (0.800) & $m_1{+}m_2$ (1.50) & $m_2/m_1$ (0.500) \\
    \midrule
    force residual & Chain (state space) & $1.078 \pm 0.26$ & $0.5367 \pm 0.13$ & $1.011 \pm 0.011$ & $0.7890 \pm 0.010$ & $1.615 \pm 0.39$ & $0.4976 \pm 0.012$ \\
     & Collider (state space) & $0.1713$\,$\ddagger$ & $0.02538$\,$\ddagger$ & $1.386$\,$\ddagger$ & $0.07589$\,$\ddagger$ & $0.1967$\,$\ddagger$ & $0.1481$\,$\ddagger$ \\
     & Collider (B-PINN) & $0.08769 \pm 0.020$ & $0.02269 \pm 0.0054$ & $0.1441 \pm 0.019$ & $0.2137 \pm 0.034$ & $0.1104 \pm 0.022$ & $0.2709 \pm 0.085$ \\
    \addlinespace
    acceleration residual & Chain (state space) & $1.021 \pm 0.30$ & $0.5062 \pm 0.15$ & $1.011 \pm 0.012$ & $0.7926 \pm 0.011$ & $1.527 \pm 0.45$ & $0.4958 \pm 0.011$ \\
     & Collider (state space) & $1.021 \pm 0.30$ & $0.5062 \pm 0.15$ & $1.011 \pm 0.012$ & $0.7926 \pm 0.011$ & $1.527 \pm 0.45$ & $0.4958 \pm 0.011$ \\
     & Collider (B-PINN) & $1.004$\,$\ddagger$ & $0.4715$\,$\ddagger$ & $1.023$\,$\ddagger$ & $0.7808$\,$\ddagger$ & $1.476$\,$\ddagger$ & $0.4699$\,$\ddagger$ \\
    \bottomrule
    \end{tabular}%
    }
    \end{small}
    \end{center}
\end{table}

\begin{table}[t]
    \caption{\textbf{Relativistic particle.} The three inference legs under both residual metrics, on one dataset and one prior, with $\sigma_r$ held at its true value. Posterior mean $\pm$ standard deviation over 14000 draws (state space) and 24000 draws (B-PINN). $\dagger$: a walk at the Euler map's stability boundary, not a posterior; its median is quoted; $\ddagger$: chains that have not merged ($\hat{R} > 1.05$), likewise a median rather than a posterior.}
    \label{tab:grid-rp-sigfix}
    \vskip 0.05in
    \begin{center}
    \begin{small}
    \resizebox{\columnwidth}{!}{%
    \begin{tabular}{@{}llrrrr@{}}
    \toprule
    \textbf{residual} & \textbf{leg} & $m_0$ (1.00) & $q$ (1.00) & $c$ (1.00) & $q/m_0$ (1.00) \\
    \midrule
    force residual & Chain (state space) & $0.8233 \pm 0.15$ & $0.8681 \pm 0.15$ & $0.8939 \pm 0.12$ & $1.057 \pm 0.055$ \\
     & Collider (state space) & $2.1{\cdot}10^{-5}$\,$\ddagger$ & $0.00383$\,$\ddagger$ & $4133$\,$\ddagger$ & $182.8$\,$\ddagger$ \\
     & Collider (B-PINN) & $0.1630 \pm 0.036$ & $0.1485 \pm 0.032$ & $1.608 \pm 0.27$ & $0.9127 \pm 0.036$ \\
    \addlinespace
    acceleration residual & Chain (state space) & $0.9941 \pm 0.29$ & $1.049 \pm 0.31$ & $0.8881 \pm 0.087$ & $1.056 \pm 0.046$ \\
     & Collider (state space) & $0.9941 \pm 0.29$ & $1.049 \pm 0.31$ & $0.8881 \pm 0.087$ & $1.056 \pm 0.046$ \\
     & Collider (B-PINN) & $1.101 \pm 0.33$ & $0.9872 \pm 0.29$ & $1.783 \pm 0.25$ & $0.8969 \pm 0.015$ \\
    \bottomrule
    \end{tabular}%
    }
    \end{small}
    \end{center}
\end{table}

%% file: FixPINNs.bib
@article{Raissi19_PINN,
  author  = {Raissi, Maziar and Perdikaris, Paris and Karniadakis, George Em},
  title   = {Physics-informed neural networks: A deep learning framework for solving forward and inverse problems involving nonlinear partial differential equations},
  journal = {Journal of Computational Physics},
  volume  = {378},
  pages   = {686--707},
  year    = {2019},
  doi     = {10.1016/j.jcp.2018.10.045}
}

@article{Wang2021_FourierFeature,
  author  = {Wang, Sifan and Wang, Hanwen and Perdikaris, Paris},
  title   = {On the eigenvector bias of {Fourier} feature networks: From regression to solving multi-scale {PDEs} with physics-informed neural networks},
  journal = {Computer Methods in Applied Mechanics and Engineering},
  volume  = {384},
  pages   = {113938},
  year    = {2021},
  doi     = {10.1016/j.cma.2021.113938}
}

@article{Yang2021_BPINN,
  author  = {Yang, Liu and Meng, Xuhui and Karniadakis, George Em},
  title   = {{B-PINNs}: {Bayesian} physics-informed neural networks for forward and inverse {PDE} problems with noisy data},
  journal = {Journal of Computational Physics},
  volume  = {425},
  pages   = {109913},
  year    = {2021},
  doi     = {10.1016/j.jcp.2020.109913}
}

@article{LiuChen1998_SMC,
  author  = {Liu, Jun S. and Chen, Rong},
  title   = {Sequential {Monte Carlo} Methods for Dynamic Systems},
  journal = {Journal of the American Statistical Association},
  volume  = {93},
  number  = {443},
  pages   = {1032--1044},
  year    = {1998},
  url     = {http://www.jstor.org/stable/2669847}
}

@inproceedings{Archambeau2007_VI,
  author    = {Archambeau, C\'{e}dric and Opper, Manfred and Shen, Yuan and Cornford, Dan and Shawe-Taylor, John},
  title     = {Variational Inference for Diffusion Processes},
  booktitle = {Advances in Neural Information Processing Systems},
  editor    = {Platt, J. and Koller, D. and Singer, Y. and Roweis, S.},
  volume    = {20},
  publisher = {Curran Associates, Inc.},
  year      = {2007},
  url       = {https://proceedings.neurips.cc/paper_files/paper/2007/file/818f4654ed39a1c147d1e51a00ffb4cb-Paper.pdf}
}

@inproceedings{Stock2023_BPINN,
  author    = {Stock, Simon and Stiasny, Jochen and Babazadeh, Davood and Becker, Christian and Chatzivasileiadis, Spyros},
  title     = {{Bayesian} Physics-Informed Neural Networks for Robust System Identification of Power Systems},
  booktitle = {2023 {IEEE} {Belgrade} {PowerTech}},
  pages     = {1--6},
  publisher = {IEEE},
  year      = {2023},
  doi       = {10.1109/PowerTech55446.2023.10202692}
}

@article{Sun2020_BPINN,
  author  = {Sun, Luning and Wang, Jian-Xun},
  title   = {Physics-constrained {Bayesian} neural network for fluid flow reconstruction with sparse and noisy data},
  journal = {Theoretical and Applied Mechanics Letters},
  volume  = {10},
  number  = {3},
  pages   = {161--169},
  year    = {2020},
  doi     = {10.1016/j.taml.2020.01.031}
}

@article{Linka2022_BPINN,
  author  = {Linka, Kevin and Sch{\"a}fer, Amelie and Meng, Xuhui and Zou, Zongren and Karniadakis, George Em and Kuhl, Ellen},
  title   = {{Bayesian} Physics-Informed Neural Networks for real-world nonlinear dynamical systems},
  journal = {Computer Methods in Applied Mechanics and Engineering},
  volume  = {402},
  pages   = {115346},
  year    = {2022},
  doi     = {10.1016/j.cma.2022.115346}
}

@article{Pensoneault2024,
  author  = {Pensoneault, Andrew and Zhu, Xueyu},
  title   = {Efficient {Bayesian} Physics Informed Neural Networks for inverse problems via Ensemble {Kalman} Inversion},
  journal = {Journal of Computational Physics},
  volume  = {508},
  pages   = {113006},
  year    = {2024},
  doi     = {10.1016/j.jcp.2024.113006}
}

@misc{Moon2025_ThermalBPINN,
  author = {Moon, Hyeonbin and Cho, Hanbin and Demeke, Wabi and Ryu, Byungki and Ryu, Seunghwa},
  title  = {Thermal Conductivity Estimation of Thermoelectric Materials with Uncertainty Quantification Using {Bayesian} Physics-Informed Neural Networks},
  note   = {arXiv:2510.16723},
  year   = {2025}
}

@misc{Shukla2026_TurbulentUQ,
  author = {Shukla, Khemraj and Zou, Zongren and Kaeufer, Theo and Triantafyllou, Michael and Karniadakis, George Em},
  title  = {Uncertainty Quantification in {PINNs} for Turbulent Flows: {Bayesian} Inference and Repulsive Ensembles},
  note   = {arXiv:2604.17156},
  year   = {2026}
}

@article{Bilionis1,
  author  = {Alberts, Alex and Bilionis, Ilias},
  title   = {Physics-informed information field theory for modeling physical systems with uncertainty quantification},
  journal = {Journal of Computational Physics},
  volume  = {486},
  pages   = {112100},
  year    = {2023},
  doi     = {10.1016/j.jcp.2023.112100}
}

@article{Bilionis2,
  author  = {Hao, Kairui and Bilionis, Ilias},
  title   = {An information field theory approach to {Bayesian} state and parameter estimation in dynamical systems},
  journal = {Journal of Computational Physics},
  volume  = {512},
  pages   = {113139},
  year    = {2024},
  doi     = {10.1016/j.jcp.2024.113139}
}

@article{Bilionis3,
  author  = {Hao, Kairui and Bilionis, Ilias},
  title   = {Neural information field filter},
  journal = {Mechanical Systems and Signal Processing},
  volume  = {226},
  pages   = {112253},
  year    = {2025},
  doi     = {10.1016/j.ymssp.2024.112253}
}

@article{Hoffman2014_NUTS,
  author  = {Hoffman, Matthew D. and Gelman, Andrew},
  title   = {The {No-U-Turn} Sampler: Adaptively Setting Path Lengths in {Hamiltonian} {Monte Carlo}},
  journal = {Journal of Machine Learning Research},
  volume  = {15},
  number  = {47},
  pages   = {1593--1623},
  year    = {2014},
  url     = {http://jmlr.org/papers/v15/hoffman14a.html}
}

@inproceedings{Murray2006,
  author    = {Murray, Iain and Ghahramani, Zoubin and MacKay, David J. C.},
  title     = {{MCMC} for doubly-intractable distributions},
  booktitle = {Proceedings of the Twenty-Second Conference on Uncertainty in Artificial Intelligence},
  pages     = {359--366},
  publisher = {AUAI Press},
  year      = {2006}
}

@article{Andrieu2010_PMMH,
  author  = {Andrieu, Christophe and Doucet, Arnaud and Holenstein, Roman},
  title   = {Particle {Markov} chain {Monte Carlo} methods},
  journal = {Journal of the Royal Statistical Society: Series B (Statistical Methodology)},
  volume  = {72},
  number  = {3},
  pages   = {269--342},
  year    = {2010},
  doi     = {10.1111/j.1467-9868.2009.00736.x}
}

@article{Gordon1993_Bootstrap,
  author  = {Gordon, Neil J. and Salmond, David J. and Smith, Adrian F. M.},
  title   = {Novel approach to nonlinear/non-{Gaussian} {Bayesian} state estimation},
  journal = {IEE Proceedings F (Radar and Signal Processing)},
  volume  = {140},
  number  = {2},
  pages   = {107--113},
  year    = {1993},
  doi     = {10.1049/ip-f-2.1993.0015}
}

@book{koller2009probabilistic,
  author    = {Koller, Daphne and Friedman, Nir},
  title     = {Probabilistic Graphical Models: Principles and Techniques},
  publisher = {MIT Press},
  address   = {Cambridge, MA},
  year      = {2009}
}

@book{goldstein2002classical,
  author    = {Goldstein, Herbert and Poole, Charles and Safko, John},
  title     = {Classical Mechanics},
  edition   = {Third},
  publisher = {Addison Wesley},
  year      = {2002}
}

@book{murray1994mathematical,
  author    = {Murray, Richard M. and Li, Zexiang and Sastry, S. Shankar},
  title     = {A Mathematical Introduction to Robotic Manipulation},
  publisher = {CRC Press},
  year      = {1994}
}

@article{Durr1978_OM,
  author  = {D{\"u}rr, Detlef and Bach, Alexander},
  title   = {The {Onsager--Machlup} Function as {Lagrangian} for the Most Probable Path of a Diffusion Process},
  journal = {Communications in Mathematical Physics},
  volume  = {60},
  number  = {2},
  pages   = {153--170},
  year    = {1978},
  doi     = {10.1007/BF01609446}
}

@article{OnsagerMachlup1953,
  author  = {Onsager, Lars and Machlup, Stefan},
  title   = {Fluctuations and Irreversible Processes},
  journal = {Physical Review},
  volume  = {91},
  number  = {6},
  pages   = {1505--1512},
  year    = {1953},
  doi     = {10.1103/PhysRev.91.1505}
}

@article{WongZakai1965,
  author  = {Wong, Eugene and Zakai, Moshe},
  title   = {On the Convergence of Ordinary Integrals to Stochastic Integrals},
  journal = {The Annals of Mathematical Statistics},
  volume  = {36},
  number  = {5},
  pages   = {1560--1564},
  year    = {1965},
  doi     = {10.1214/aoms/1177699916}
}

@book{Sarkka2019,
  author    = {S{\"a}rkk{\"a}, Simo and Solin, Arno},
  title     = {Applied Stochastic Differential Equations},
  series    = {Institute of Mathematical Statistics Textbooks},
  volume    = {10},
  publisher = {Cambridge University Press},
  year      = {2019},
  doi       = {10.1017/9781108186735}
}

@misc{jax2018github,
  author = {Bradbury, James and Frostig, Roy and Hawkins, Peter and Johnson, Matthew James and Leary, Chris and Maclaurin, Dougal and Necula, George and Paszke, Adam and Vander{P}las, Jake and Wanderman-{M}ilne, Skye and Zhang, Qiao},
  title  = {{JAX}: composable transformations of {Python+NumPy} programs},
  howpublished = {\url{https://github.com/jax-ml/jax}},
  year   = {2018}
}

@misc{phan2019numpyro,
  author = {Phan, Du and Pradhan, Neeraj and Jankowiak, Martin},
  title  = {Composable Effects for Flexible and Accelerated Probabilistic Programming in {NumPyro}},
  note   = {arXiv:1912.11554},
  year   = {2019}
}

@article{bingham2019pyro,
  author  = {Bingham, Eli and Chen, Jonathan P. and Jankowiak, Martin and Obermeyer, Fritz and Pradhan, Neeraj and Karaletsos, Theofanis and Singh, Rohit and Szerlip, Paul A. and Horsfall, Paul and Goodman, Noah D.},
  title   = {{Pyro}: Deep Universal Probabilistic Programming},
  journal = {Journal of Machine Learning Research},
  volume  = {20},
  number  = {28},
  pages   = {1--6},
  year    = {2019}
}

@misc{Bolt2026_NIFFReplication,
  author       = {Bolt, Maxwell},
  title        = {niff-replication: A {JAX} reimplementation of the {Neural Information Field Filter}},
  howpublished = {\url{https://github.com/BoltMaxwell/niff-replication}},
  note         = {Commit \texttt{dab4403}},
  year         = {2026}
}
